\documentclass[letterpaper, 10 pt, conference]{ieeeconf}
\usepackage{tabularx}
\usepackage{booktabs}
\usepackage{multirow}
\usepackage{amsmath}
\usepackage{amssymb}
\usepackage[noadjust]{cite}
\usepackage{xparse} 
\usepackage{amsfonts} 
\usepackage{bbm}
\usepackage[svgnames]{xcolor}
\usepackage{mathtools}
\usepackage{graphicx}
\usepackage{subcaption}
\usepackage{calc}
\usepackage{balance}
\usepackage{url}
\usepackage{algorithm}
\usepackage{algorithmic}
\usepackage{wrapfig}
\usepackage{ragged2e}

\makeatletter
\let\NAT@parse\undefined
\makeatother
\usepackage[colorlinks=true,linkcolor=DarkBlue,citecolor=DarkBlue]{hyperref}
\usepackage[capitalize]{cleveref}

\newcolumntype{Y}{>{\centering\arraybackslash}X}

\usepackage{pifont}%
\newcommand{\cmark}{\ding{51}}%
\newcommand{\xmark}{\ding{55}}%

\newcommand\hl[1]{%
  \bgroup
  \hskip0pt\color{blue!90!black}%
  #1%
  \egroup
}

\newif\ifhighlight

\ifhighlight
\else
    \renewcommand\hl[1]{#1}
\fi

\usepackage{multibib}
\newcites{app}{References}  %

\usepackage{xparse}

\NewDocumentCommand\bbm{}{ \begin{bmatrix} }
\NewDocumentCommand\ebm{}{ \end{bmatrix} }

\NewDocumentCommand\Real{}{ \mathbb{R} }

\newcommand{\mdpm}{\mathcal{M}}
\newcommand{\mdps}{\mathcal{S}}
\newcommand{\mdpa}{\mathcal{A}}

\newcommand{\mdpp}{\mathcal{P}}
\newcommand{\mdprho}{\rho_0}

\newcommand{\buffer}{\mathcal{B}}
\newcommand{\tasks}{\mathcal{T}}

\NewDocumentCommand\ExpectationSamp{mm}{ \mathbb{E}_{#1}\left[#2\right] }

\DeclareMathOperator*{\argmax}{arg\,max}

\IEEEoverridecommandlockouts                              %
\pdfoutput=1

\title{\LARGE \bf Efficient Imitation Without Demonstrations via\\ Value-Penalized Auxiliary Control from Examples}

\author{Trevor Ablett$^{1}$, Bryan Chan$^{2}$, Jayce Haoran Wang$^{1}$, and Jonathan Kelly$^{1}$
\thanks{$^{1}$Authors are with the Space \& Terrestrial Autonomous Robotic Systems (STARS) Laboratory at the University of Toronto Institute for Aerospace Studies (UTIAS), Toronto, Ontario, Canada, M3H~5T6. Email: {\tt\small <first name>.<last name>@robotics.utias.utoronto.ca}}
\thanks{$^{2}$Author is with the Department of Computing Science at the University of Alberta, Edmonton, Alberta, Canada, T6G 2E8. Email: {\tt\small bryan.chan@ualberta.ca}}
}

\begin{document}

\maketitle
\thispagestyle{empty}
\pagestyle{empty}

\begin{abstract}
\label{sec:abstract}
Common approaches to providing feedback in reinforcement learning are the use of hand-crafted rewards or full-trajectory expert demonstrations.
Alternatively, one can use examples of completed tasks, but such an approach can be extremely sample inefficient.
We introduce value-penalized auxiliary control from examples (VPACE), an algorithm that significantly improves exploration in example-based control by adding examples of simple auxiliary tasks and an above-success-level value penalty. 
Across both simulated and real robotic environments, we show that our approach substantially improves learning efficiency for challenging tasks, while maintaining bounded value estimates.
Preliminary results also suggest that VPACE may learn more efficiently than the more common approaches of using full trajectories or true sparse rewards.
Project site: \url{https://papers.starslab.ca/vpace/}.
\end{abstract}

\section{Introduction}
\label{sec:introduction}

Example-based control (EBC) is a form of reinforcement learning in which example states of completed tasks are the only form of feedback \cite{eysenbachReplacingRewardsExamples2021}.
Obtaining example states can be far less laborious than designing a reward function or gathering expert trajectories, and furthermore removes any possibility of using suboptimal expert trajectories.
However, just as in the case of sparse rewards, excluding information on how goal states are reached can lead to highly inefficient learning (e.g., an example of a loaded dishwasher provides no information about the long sequence of actions required to complete the task).
\textbf{\textit{Can we improve the sample efficiency of example-based control?}}

\begin{figure}
    \centering
    \includegraphics[width=\linewidth]{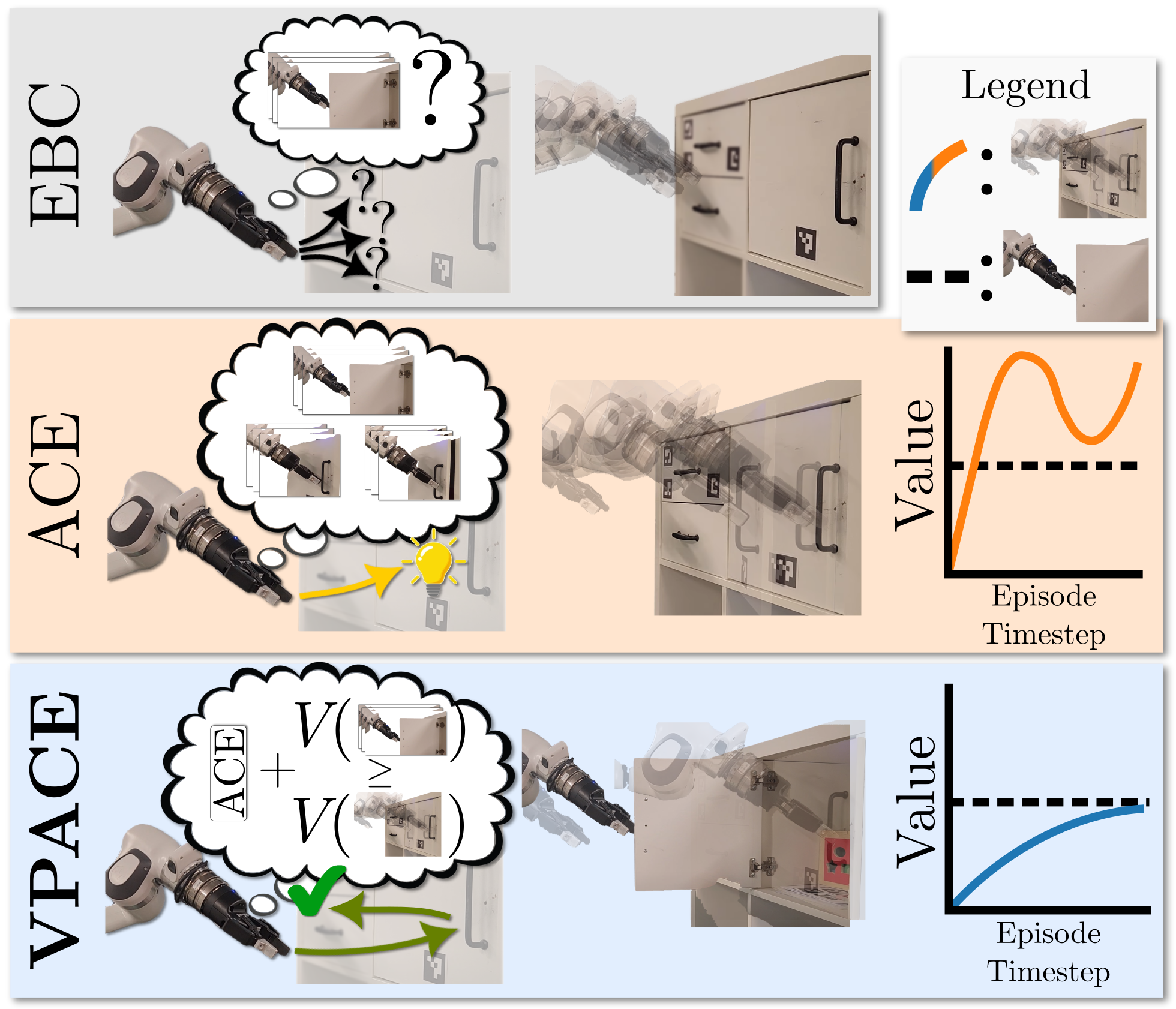}
    \caption{
        Example-based control (EBC) is inefficient due to poor exploration resulting from the inherently sparse nature of individual examples of success as a means for feedback.
        Auxiliary control from examples (ACE) remedies this by adding scheduled exploration of semantically meaningful auxiliary tasks, but can result in poor performance due to the unbounded value error of highly exploratory data.
        We combine ACE with value penalization (VP) as VPACE to efficiently learn from examples of success.
    }
    \label{fig:vpace_motivation}
    \vspace{-12pt}
\end{figure}

To answer this question, we propose a new example-based control method, \textbf{v}alue-\textbf{p}enalized \textbf{a}uxiliary \textbf{c}ontrol from \textbf{e}xamples (\textbf{VPACE}).
Since hierarchical RL (HRL) has shown success in improving exploration in robotics domains \cite{nachumWhyDoesHierarchy2019},
we leverage an HRL approach known as scheduled auxiliary control (SAC-X) \cite{riedmillerLearningPlayingSolving2018} to improve exploration in EBC.
The SAC-X framework enables practitioners to introduce a set of simple and often reusable auxiliary tasks, in addition to the main task, that help the agent to explore the environment.
In this work, the full set of auxiliary tasks that we use across all main tasks are reach, grasp, lift, and release.
Instead of defining main and auxiliary tasks with sparse rewards \cite{riedmillerLearningPlayingSolving2018} or full expert trajectories \cite{ablettLearningGuidedPlay2023}, we define tasks with examples.
For each auxiliary task there is a corresponding auxiliary policy that learns to match the set of examples.
A scheduler periodically chooses and executes the different auxiliary policies, in addition to the main policy, generating a more diverse set of data to learn from.

We find that the na\"{i}ve application of SAC-X to EBC can result in overestimated values, leading to sample inefficiency and poor performance.
To remedy this, and show how SAC-X can be effectively applied to EBC, this work makes the following contributions:
(i) we demonstrate that the introduction of the SAC-X framework significantly improves exploration and learning efficiency in EBC,
(ii) we remedy this overestimation problem by introducing a value-penalization method that is based on the expected value of the examples,
(iii) we conduct experiments across four environments with 19 simulated and two real tasks to show the improved sample efficiency and final performance of VPACE over EBC, inverse reinforcement learning, and an exploration bonus.
(iv) We compare to the use of full trajectories and true sparse rewards, and observe that VPACE has higher sample efficiency than both.

\section{Related Work}
\label{sec:related-work}

\begin{figure}
	\centering
	\includegraphics[width=0.8\linewidth]{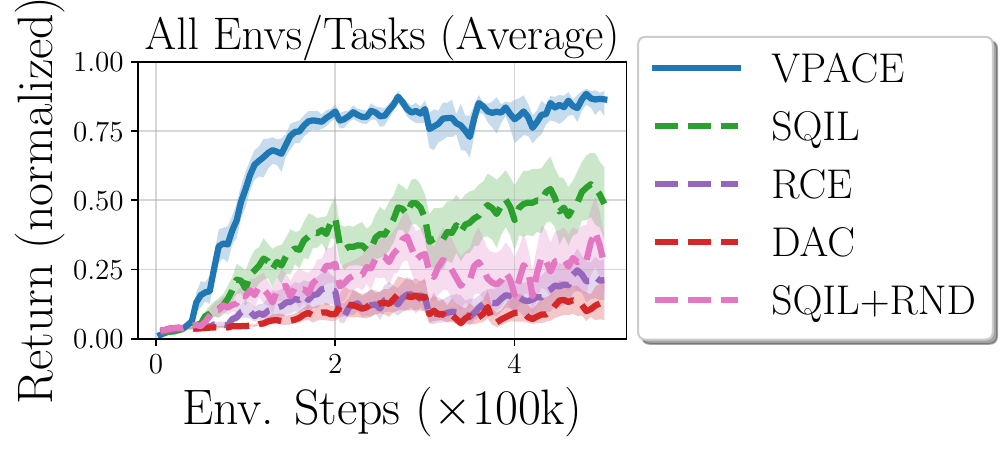}
	\caption{
		Average performance across all environments and tasks studied in this work.
		Results are shown as an interquartile mean with five seeds per algorithm and task, and shaded regions show 95\% stratified bootstrap confidence intervals \cite{agarwalDeepReinforcementLearning2021}.
	}
	\label{fig:perf_results_all_envs_avg}
	\vspace{-12pt}
\end{figure}

Sparse rewards are a desirable form of feedback for learning unbiased, optimal policies in reinforcement learning (RL), but they are not always obtainable, and present an immense exploration challenge on long-horizon tasks \cite{guptaUnpackingRewardShaping2022}.
Reward shaping \cite{ngPolicyInvarianceReward1999} and dense rewards can help alleviate the exploration problem in robotics \cite{popovDataefficientDeepReinforcement2017,yuMetaWorldBenchmarkEvaluation2019}, but designing dense rewards is difficult for practitioners \cite{andrychowiczHindsightExperienceReplay2017}. %
An alternative to manually-defined rewards is to perform inverse RL (IRL), in which a reward function is recovered from demonstrations, and a policy is learned either subsequently \cite{ngAlgorithmsInverseReinforcement2000} or simultaneously via adversarial imitation learning (AIL) \cite{kostrikovDiscriminatorActorCriticAddressingSample2019,reddySQILImitationLearning2020,hoGenerativeAdversarialImitation2016,fuLearningRobustRewards2018}.

Like dense rewards, full trajectory demonstrations can be hard to acquire, suboptimal, or biased.
Unlike IRL/AIL, in \textit{example-based control} (EBC), a learning agent is only provided distributions of single \textit{successful example states}.
Previous EBC approaches include using generative AIL (GAIL, \cite{hoGenerativeAdversarialImitation2016}) directly (VICE, \cite{fuVariationalInverseControl2018}), soft actor critic (SAC, \cite{haarnojaSoftActorCriticOffPolicy2018}) with an additional mechanism for generating extra success examples (VICE-RAQ, \cite{singhEndtoEndRoboticReinforcement2019}), performing offline RL with conservative Q-learning (CQL, \cite{kumarConservativeQLearningOffline2020}) and a learned reward function \cite{hatchContrastiveExampleBasedControl2023}, and using SAC with a classifier-based reward (RCE, \cite{eysenbachReplacingRewardsExamples2021}).

All EBC methods can naturally suffer from poor exploration, given that success examples are akin to sparse rewards.
Hierarchical reinforcement learning (HRL) aims to leverage multiple levels of abstraction in long-horizon tasks \cite{suttonMDPsSemiMDPsFramework1999}, improving exploration in RL \cite{robertSampleComplexityGoalConditioned2023,nachumWhyDoesHierarchy2019,nairOvercomingExplorationReinforcement2018}.
Scheduled auxiliary control (SAC-X, \cite{riedmillerLearningPlayingSolving2018}) combines a scheduler with semantically meaningful and simple auxiliary sparse rewards or auxiliary full expert trajectories (LfGP, \cite{ablettLearningGuidedPlay2023, ablettLearningGuidedPlay2021, xiangSCAIRLShareCriticAdversarial2024}).
Like us, \cite{wuExampleDrivenModelBasedReinforcement2021} combined EBC with hierarchical learning, but their approach required a symbolic planner at test time, and generated very slow policies: our policies are fast and reactive, and don't require high-level planning at test time.

Learning value functions in the off-policy setting can be challenging due to the deadly triad \cite{suttonReinforcementLearningIntroduction2018}.
Regularization and clipping techniques have been applied to address various problems such as stabilizing the bootstrapping target \cite{andrychowiczHindsightExperienceReplay2017,adamczykBoostingSoftQLearning2024} and preventing overfitting and out-of-distribution samples \cite{kumarConservativeQLearningOffline2020,jamesCoarsetoFineQAttentionEfficient2022}.
Our proposed value-penalization technique specifically targets overestimation in EBC.

\section{Example-Based Control with Value-Penalization and Auxiliary Tasks}
\label{sec:method_main}

Our goal is to generate an agent that can complete a task given, as opposed to rewards or demonstrations, only \textit{final-state examples} of a successfully completed task, with as few environment interactions as possible.
We also assume access to final state examples of a small set of reusable auxiliary tasks.
We begin by formally describing the problem setting for example-based control in \cref{subsec:problem}.
In \cref{subsec:method_learning_an_agent}, we describe how scheduled auxiliary tasks can be applied to example-based control.
Finally, motivated by the increased exploration diversity of the multitask framework, we propose a new Q-estimation objective in \cref{subsec:value_pen} that leverages value penalization for improved learning stability.

\begin{table}
	\centering
	\small
	\caption{VPACE fills a gap in literature.}
	\begin{tabular}{lll}
		&  Single-task   & Sched. Aux. Ctrl  \\
		\midrule
		Dense Reward   & RL             & \cite{riedmillerLearningPlayingSolving2018}     \\
		Sparse Reward  & \cite{ngPolicyInvarianceReward1999,andrychowiczHindsightExperienceReplay2017}            & \cite{riedmillerLearningPlayingSolving2018}     \\
		Dense Expert   & IRL            & \cite{ablettLearningGuidedPlay2021,ablettLearningGuidedPlay2023,xiangSCAIRLShareCriticAdversarial2024} \\
		Sparse Expert  & \cite{eysenbachReplacingRewardsExamples2021,fuVariationalInverseControl2018,hatchContrastiveExampleBasedControl2023}     & \textbf{VPACE} \\
	\end{tabular}
	\label{tab:vpace_gap}
	\vspace{-12pt}
\end{table}

\subsection{Problem Setting}
\label{subsec:problem}

A Markov decision process (MDP) is defined as $\mdpm = \langle \mdps, \mdpa, R, \mdpp, \mdprho, \gamma \rangle$, where the sets $\mdps$ and $\mdpa$ are respectively the state and action space, $\mdpp$ is the state-transition environment dynamics distribution, $\mdprho$ is the initial state distribution, $\gamma$ is the discount factor, and the true reward $R : \mathcal{S} \times \mathcal{A} \rightarrow \Real$ is unknown.
Actions are sampled from a stochastic policy $\pi(a|s)$.
The policy $\pi$ interacts with the environment to yield experience $\left( s_t, a_t, s_{t+1} \right)$, generated by $s_0 \sim \mdprho(\cdot), a \sim \pi(\cdot | s_t)$, and $s_{t+1} \sim \mdpp(\cdot | s_t, a_t)$.
The gathered experience $\left( s_t, a_t, s_{t+1} \right)$ is then stored in a buffer $\buffer$, which may be used throughout learning.
For any variables $x_t, x_{t+1}$, we may drop the subscripts and use $x, x'$ instead when the context is clear.

In this work, we focus on \textit{example-based control} (EBC), a more difficult form of imitation learning where we have no access to rewards or demonstrations, but instead are given a set of example states of a completed task: $s^* \in \buffer^*$, where $ \buffer^* \subseteq \mathcal{S}$ and $\lvert \buffer^* \rvert < \infty$.
The goal is to (i) leverage $\buffer^*$ and $\buffer$ to learn or define a state-conditional reward function $\hat R: \mathcal{S} \to \mathbb{R}$ that satisfies $\hat R(s^*) \geq \hat R(s)$ for all $(s^*, s) \in \buffer^* \times \buffer$, and (ii) learn a policy $\hat\pi$ that maximizes the expected return $\hat\pi = \argmax_{\pi} \mathbb{E}_\pi\left[ \sum_{t=0}^{\infty} \gamma^{t} \hat{R}(s_t) \right]$.

For any policy $\pi$, we can define the value function and Q-function respectively to be $V^{\pi}(s) = \mathbb{E}_{\pi} \left[ Q^{\pi}(s, a) \right]$ and $Q^{\pi}(s, a) = \hat{R}(s) + \gamma \mathbb{E}_{\mdpp}\left[ V^{\pi}(s') \right]$,
corresponding to the return-to-go from state $s$ (and action $a$).
Temporal difference (TD) algorithms aim to estimate $V^\pi$ or $Q^\pi$ to evaluate a policy \cite{suttonReinforcementLearningIntroduction2018}.
Given a reward model $\hat{R}(\cdot)$, we can say $\hat{R}(s^*), s^* \in \buffer^*$, indicates reward for successful states and $\hat{R}(s), s \in \buffer$, for all other states.
Assuming that $s^*$ transitions to itself, then for policy evaluation with mean-squared error (MSE), we can write the TD targets for non-successful and successful states, $y: \mathcal{S} \times \mathcal{S} \to \mathbb{R}$, of Q-updates as
\begin{equation}
    \label{eq:target_policy_bellman}
     y(s, s') = \hat{R}(s) + \gamma \mathbb{E}_{\pi}\left[ Q(s', a') \right],
\end{equation}
\begin{equation}
    \label{eq:target_expert_bellman}
    y(s^*, s^*) = \hat{R}(s^*) + \gamma \mathbb{E}_{\pi}\left[Q(s^*, a')\right],
\end{equation}
where $(s, \cdot, s') \sim \buffer$ and $s^* \sim \buffer^*$.
The reward function $\hat{R}$ can be based on a learned discriminator that differentiates between $s \sim \buffer$ and $s^* \sim \buffer^*$, akin to a state-only version of adversarial imitation learning \cite{kostrikovDiscriminatorActorCriticAddressingSample2019,hoGenerativeAdversarialImitation2016,fuLearningRobustRewards2018}, or defined directly, such as $\hat{R}(s^*) = 1$ and $\hat{R}(s) = 0$ \cite{ eysenbachReplacingRewardsExamples2021,reddySQILImitationLearning2020}.

\subsection{Learning a Multitask Agent from Examples}
\label{subsec:method_learning_an_agent}

We alleviate the challenging exploration problem of EBC by introducing auxiliary control from examples (ACE), an application of the scheduled auxiliary control framework \cite{riedmillerLearningPlayingSolving2018, ablettLearningGuidedPlay2023}.
ACE leverages a scheduler to sequentially choose and execute individual intention policies allowing for more diverse state coverage to facilitate faster learning of the main task.
Crucially, because the method is off-policy, each intention can learn from data generated by any other intention, and all policy interaction data, regardless of which intention generated it, is stored in $\buffer$.

Formally, given an MDP $\mdpm$, a task $\tasks$ is defined by a task-specific example buffer $\buffer^*_\tasks$.
EBC methods such as RCE \cite{eysenbachReplacingRewardsExamples2021} aim to exclusively complete the main task $\tasks_{\text{main}}$,
while ACE adds auxiliary tasks $\tasks_{\text{aux}} = \left\{ \tasks_1, \dots, \tasks_K \right\}$ during learning.
We refer the set of all tasks as $\tasks_{\text{all}} = \tasks_\text{aux} \cup \left\{ \tasks_\text{main} \right\}$.
ACE agents are composed of two types of policies---intentions for each task and a scheduler.

\subsubsection{Intentions}
\label{subsubsec:intentions}
For each task $\tasks \in \tasks_{\text{all}}$, the corresponding intention consists of a task-specific policy $\pi_\tasks$, Q-function $Q_\tasks$, and state-conditioned reward $\hat{R}_\tasks$.
ACE optimizes the task-specific policies by maximizing the policy optimization objective
\begin{equation}
    \label{eq:multi_pol_obj}
    \mathcal{L}(\pi; \tasks) = \mathbb{E}_{\buffer, \pi_\tasks}\left[ Q_\tasks(s, a) \right].
\end{equation}
The task-specific Q-functions are optimized via minimization of the Bellman residual
\begin{align}
\begin{split}
    \label{eq:multi_q_obj}
    \mathcal{L}(Q; \tasks) = &\mathbb{E}_{\buffer, \pi_\tasks}\left[ (Q_\tasks(s, a) - y_\tasks(s, s'))^2 \right]\\
    &+ \mathbb{E}_{\buffer^*_\tasks, \pi_\tasks}\left[ (Q_\tasks(s^*, a) - y_\tasks(s^*, s^*))^2 \right],
\end{split}
\end{align}
where $y_\tasks$ are TD targets defined based on \cref{eq:target_policy_bellman,eq:target_expert_bellman} with task-specific reward $\hat{R}_\tasks$.
Intuitively, each $\pi_\tasks$ aims to maximize the estimated task-specific value $Q_\tasks$.

\subsubsection{Scheduler}
\label{subsubsec:scheduler}

\begin{figure}
    \centering
    \includegraphics[width=\linewidth]{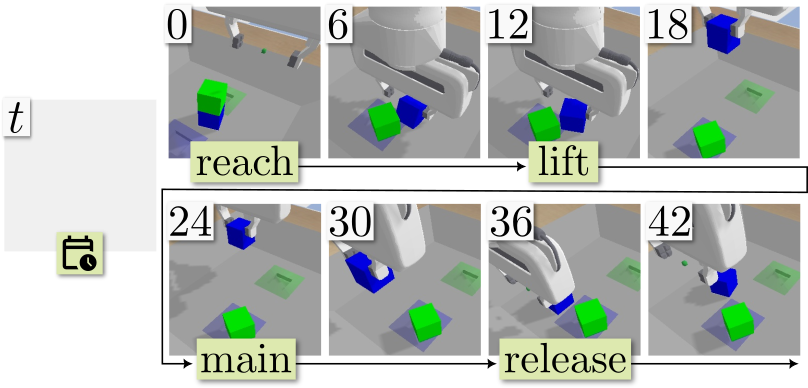}
    \caption{
        An example of fixed-period scheduler choices throughout an \texttt{Unstack-Stack} exploratory episode.
    }
    \label{fig:explore_examples}
    \vspace{-12pt}
\end{figure}

Given the $K + 1$ intentions, a scheduler periodically selects a policy $\pi_\tasks$ to execute within an episode.
Since each $\pi_\tasks$ aims to solve its own task, one can expect the transitions gathered composing by different $\pi_\tasks$ within the same episode to be diverse.
In practice, we implement the scheduler to have a fixed period, guaranteeing a specific number of policy switches within each episode, depending on the episode horizon.
We use a weighted random scheduler (WRS) with hyperparameter $p_{\tasks_\text{main}}$ where the probability of choosing the main task or an auxiliary task is $p_{\tasks_\text{main}}$ and $p_{\tasks_k} = (1 - p_{\tasks_\text{main}})/K$, respectively.
We combine a WRS with a small set of simple handcrafted high-level trajectories (e.g., \textit{reach} then \textit{grasp} then \textit{lift}).
The handcrafted trajectory definitions are reusable between main tasks, and are not required to use our framework.
\cite{ablettLearningGuidedPlay2023} demonstrated that this approach performed better than a more complex learned scheduler.
At test time, the scheduler is unused, and only $\pi_{\text{main}}$ is evaluated.

\subsection{Value Penalization in Example-Based Control}
\label{subsec:value_pen}
A scheduled multitask agent exhibits far more diverse behaviour than a single-task agent \cite{riedmillerLearningPlayingSolving2018,ablettLearningGuidedPlay2023}.
We show in \cref{fig:per_ts_q_values_main,fig:per_ts_q_ood_example} that the buffer generated by this behavior, consisting of transitions resulting from multiple policies, can result in highly overestimated Q-values in EBC.
This overestimation leads the policy to maximize an incorrect objective.
In this section, we propose a novel penalty for TD algorithms that encourages Q-estimates to stay within the range of valid returns with respect to the reward model.
This penalty applies to both the single-task and multitask regime.
For simplicity, we describe value penalization for the former.

\begin{figure*}
    \centering
    \includegraphics[width=\linewidth]{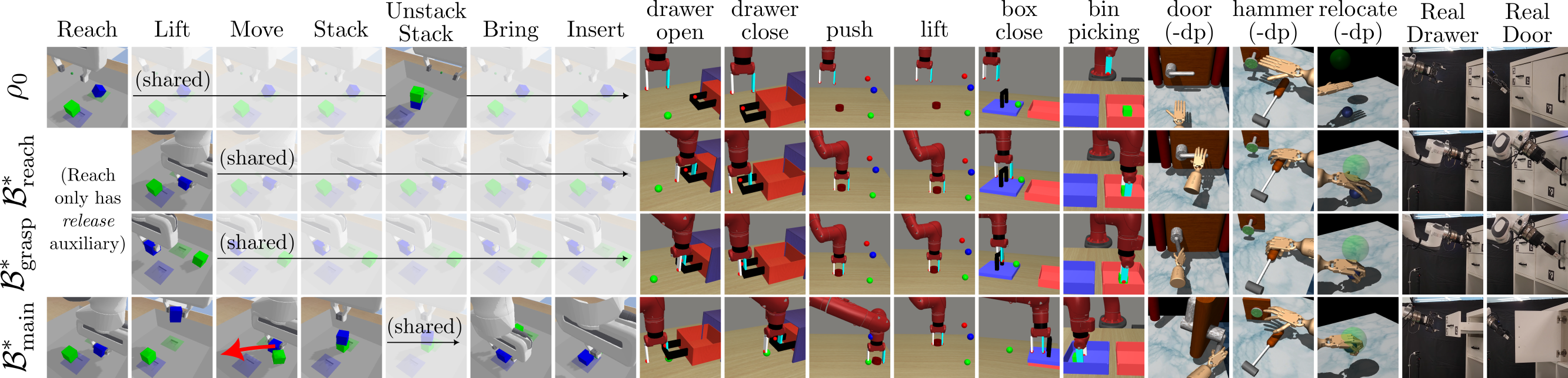}
    \caption{
        Samples from initial-state distribution $\rho_0$, auxiliary task examples $\buffer^*_\text{aux}$, and main task examples $\buffer^*_\text{main}$ for all tasks.
        The simulated Panda tasks additionally share $\buffer^*_{\text{release}}$ and $\buffer^*_{\text{lift}}$, while the Adroit original and dp environments share data since $\mdps$ does not change.
    }
    \label{fig:env_examples_main_paper}
    \vspace{-12pt}
\end{figure*}

Notice that regressing to TD targets \cref{eq:target_policy_bellman,eq:target_expert_bellman} will eventually satisfy the Bellman equation, but in the short term the TD targets do not satisfy $y(s, s') \leq y(s^*, s^*)$.
This is because the TD targets are computed by bootstrapping from a Q-estimate that may not satisfy the Bellman equation and can exceed the bounds of valid Q-values, implying that approximation error of $Q$ updated via MSE can be uncontrollable.
We resolve this issue with a penalty for our TD updates for $s \in \buffer$. %
We add a minimum and a maximum $Q^{\pi}(s,a)$ as $Q^\pi_{\text{min}} = \hat{R}_{\text{min}} / (1 - \gamma)$ and $Q^\pi_{\text{max}} = \mathbb{E}_{\buffer^*} \left[ V^\pi(s^*) \right]$, where $\hat{R}_{\text{min}} \leq \hat{R}(s)$ for all $s \in \buffer$.
Then, the value penalty is defined to be
\begin{align}
\begin{split}
\label{eq:value_penalized_q_update}
    \mathcal{L}^\pi_{\text{pen}}(Q) = \lambda \mathbb{E}_{\buffer} [ &\left( \max(Q(s,a) - Q^\pi_{\text{max}}, 0) \right)^2\\
    &+ \left( \max(Q^\pi_{\text{min}} - Q(s,a), 0) \right)^2 ],
\end{split}
\end{align}
where $\lambda \geq 0$ is a hyperparameter.
When $\lambda \to \infty$, \cref{eq:value_penalized_q_update} becomes a hard constraint. %
It immediately follows that $y(s, s') \leq y(s^*, s^*)$ holds with TD updates \cref{eq:target_policy_bellman,eq:target_expert_bellman}.
We add value penalization $\mathcal{L}^\pi_{\text{pen}}(Q)$ to the MSE loss as a regularization term for learning the Q-function.

\section{Experiments}
\label{sec:experiments}

We aim to answer the following questions through our experiments:
\textbf{(RQ1)} How does the sample efficiency of VPACE compare to EBC, inverse RL, and exploration-bonus baselines?
\textbf{(RQ2)} How important is it to include value penalization (VP) with ACE?
\textbf{(RQ3)} How does our value penalty compare to existing Q regularizers?
\textbf{(RQ4)} How does VPACE compare to algorithms that use full trajectories and true sparse rewards?

\subsection{Experimental Setup}
\label{sec:exp_env_details}

\subsubsection{Environments}
\label{subsubsec:environments}

We conduct experiments in a large variety of tasks and environments, including those originally used in LfGP \cite{ablettLearningGuidedPlay2023} and RCE \cite{eysenbachReplacingRewardsExamples2021}.
Specifically, the tasks in \cite{ablettLearningGuidedPlay2023} involve a simulated Franka Emika Panda arm, a blue and green block, a fixed ``bring" area for each block, and a small slot with $<$1 mm tolerance for inserting each block.
This environment provides various manipulation tasks that share the same state-action space.
The tasks in \cite{eysenbachReplacingRewardsExamples2021} are a modified subset of those from \cite{yuMetaWorldBenchmarkEvaluation2019}, involving a simulated Sawyer arm, and three of the Adroit hand tasks originally presented in \cite{rajeswaran*LearningComplexDexterous2018}.
We also generate three modified delta-position (dp) Adroit hand environments, because we found that policies learned in the original absolute-position environments lack finesse and exploit simulator bugs.
Finally, we study drawer and door opening tasks with a real Franka Emika Panda.

\subsubsection{Baselines}
\label{subsubsec:baselines}

\begin{figure*}[t]
    \centering
    \includegraphics[width=\textwidth]{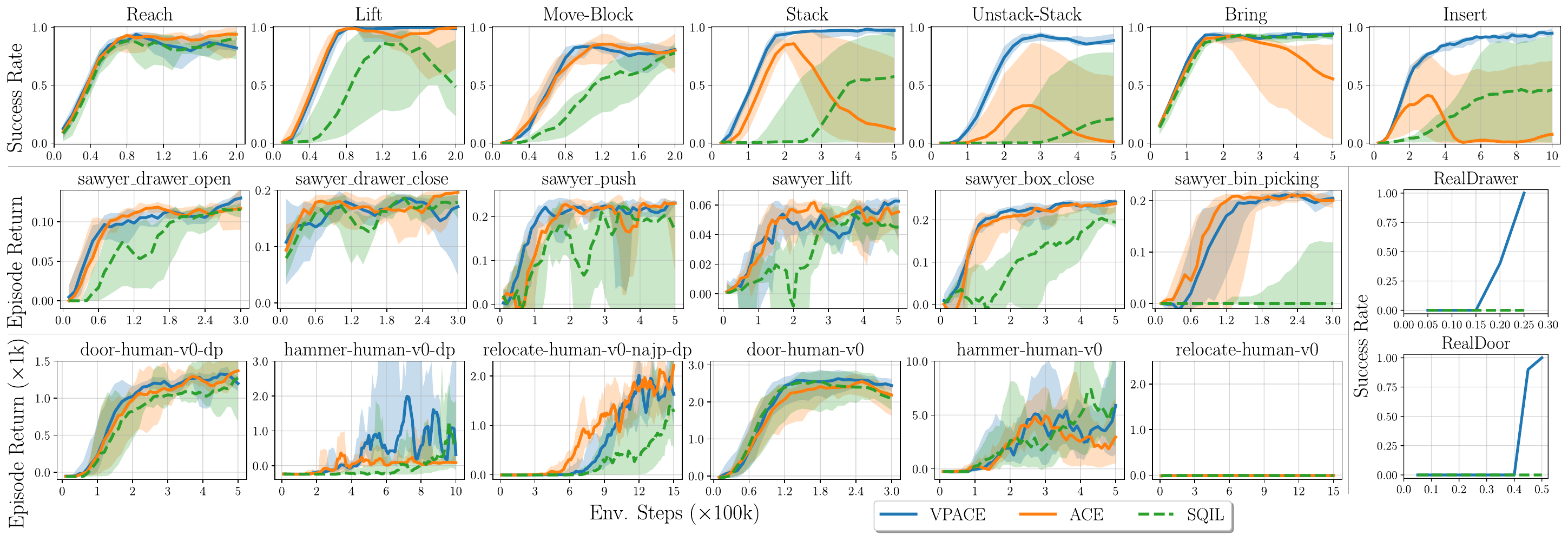}
    \caption{
        Sample efficiency performance plots for main task only for all main tasks.
        Performance is an interquartile mean (IQM) across 5 timesteps and 5 seeds with shaded regions showing 95\% stratified bootstrap confidence intervals \cite{agarwalDeepReinforcementLearning2021}.
        Other baselines (RCE, DAC, SQIL+RND) are shown only in \cref{fig:perf_results_sep_env_avgs} for clarity.
        The use of ACE significantly improves upon SQIL, and the addition of VP (VPACE) resolves instability issues that occur in \texttt{Stack}, \texttt{Unstack-Stack}, \texttt{Bring}, \texttt{Insert}, and \texttt{hammer-human-v0-dp}.
    }
    \label{fig:perf_results_all_summary}
    \vspace{-12pt}
\end{figure*}

\begin{figure}
    \centering
    \includegraphics[width=\linewidth]{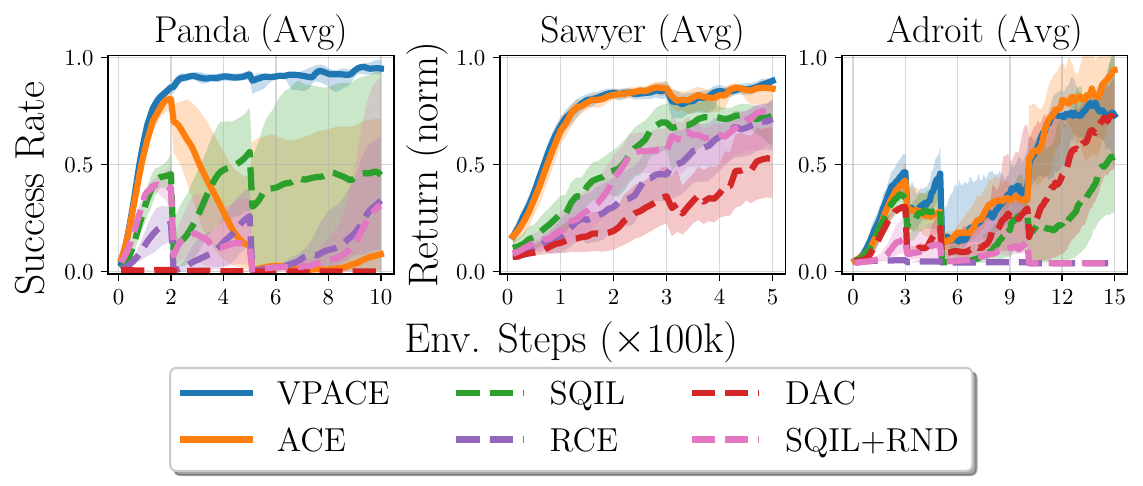}
    \caption{
        Average performance across all simulated main tasks, separated by environment, including baselines not shown in \cref{fig:perf_results_all_summary}.
    }
    \label{fig:perf_results_sep_env_avgs}
    \vspace{-12pt}
\end{figure}

We consider many baselines, including an approach based on recursively classifying examples (\textbf{RCE}, \cite{eysenbachReplacingRewardsExamples2021}), a learned reward model for off-policy RL (\textbf{DAC}, \cite{kostrikovDiscriminatorActorCriticAddressingSample2019}), a defined reward model for off-policy RL (\textbf{SQIL}, \cite{reddySQILImitationLearning2020}), and a combination of the best performing baseline (SQIL) with a method that provides an exploration bonus to unseen data (\textbf{SQIL+RND}, \cite{burdaExplorationRandomNetwork2019, ballochExplorationAllYou2024}).
Finding that SQIL significantly outperforms the baselines, we use it as the default reward model for our main experiments, although VP and ACE can be added to any of the approaches described above.
Specifically, in our results, \textbf{VPACE} refers to SQIL with both VP and ACE applied, while \textbf{ACE} refers to SQIL with ACE and \textit{without} VP applied.

\subsubsection{Implementation}
\label{subsubsec:implementation}

Observation space, action space, and auxiliary task details are shown in \cref{tab:env_details}.
All simulated tasks use 200 examples per task, while the real world tasks use 50 examples per task.
All implementations are built on LfGP \cite{ablettLearningGuidedPlay2023,rl_sandbox} and SAC \cite{haarnojaSoftActorCriticOffPolicy2018}.
For more environment and implementation details, see our open-source code.

\begin{table}
    \centering
    \scriptsize
    \caption{
        Environment details.
    }
    \begin{tabular}{p{1cm}p{2.1cm}p{2.4cm}p{1.3cm}}
        \toprule
                      & Obs. Space & Act Space & Aux. Tasks  \\
        \midrule
        \textit{Sim. Panda} \cite{ablettLearningGuidedPlay2023,ablettLearningGuidedPlay2021} & pos, vel, grip pos, prev. grip pos, obj. pos, obj. vel & XYZ delta-pos, binary grip & reach, grasp, lift, release \\
        \textit{Sawyer} \cite{eysenbachReplacingRewardsExamples2021,yuMetaWorldBenchmarkEvaluation2019} & pos, obj. pos, grip pos, 3 frame stack & XYZ delta-pos, binary grip & reach, grasp \\
        \textit{Adroit} \cite{eysenbachReplacingRewardsExamples2021,rajeswaran*LearningComplexDexterous2018} & pos, vel, obj. pos, finger pos & (up to) 6-DOF abs. / delta-pos, finger pos & reach, grasp \\
        \textit{Real Panda} & pos, obj. pos (ArUco \cite{garrido-juradoAutomaticGenerationDetection2014}), grip pos, 2 frame stack & X / XY delta-pos, binary grip & reach, grasp \\
        \bottomrule
    \end{tabular}
    \label{tab:env_details}
    \vspace{-12pt}
\end{table}

\subsection{Performance Results}

\subsubsection{Main Task VP and ACE Benefits}
\label{subsec:main_task_performance}

To answer RQ1, we compare VPACE with existing EBC, inverse RL, and exploration-bonus approaches on all tasks and evaluate their success rates.
Since the Sawyer and Adroit tasks do not evaluate success, we only report returns.
Policies are evaluated at 25k (environment step) intervals for 50 episodes for the simulated Panda tasks, 10k intervals for 30 episodes for the Sawyer and Adroit tasks, and 5k intervals for 10 episodes for the real Panda tasks.
\cref{fig:perf_results_all_summary} shows that VPACE has significant improvement over other approaches in both sample efficiency and final performance, particularly for the most difficult tasks, such as \texttt{Unstack-Stack} and \texttt{Insert}.
We can observe that VPACE can consistently solve all tasks, while other EBC methods have significantly wider confidence intervals, especially in the Panda environment.
Another notable observation is that SQIL+RND is unable to improve upon SQIL.
We speculate that the exploration bonus term actually diverts the policy from solving the main task in order to maximize the intrinsic return.

\subsubsection{ACE without VP}
\label{subsubsec:ace_without_vp}

We also include ACE in \cref{fig:perf_results_all_summary} to address RQ2.
In the simulated Panda environment, ACE performance significantly degrades as training continues.
As elaborated on in \cref{sec:val_pen_and_cal_of_qvalues}, we hypothesize that this is correlated to value overestimation being exacerbated by executing multiple policies.
For Sawyer and Adroit environments, VP does not provide benefit (i.e., VPACE and ACE have similar performance), which might be due to the tasks being simpler and having smaller observation spaces than the Panda tasks.
Nonetheless, VP does not harm performance and including auxiliary tasks provides a clear benefit.

\subsubsection{Real Robot Performance}
\label{subsubsec:real_robot_performance}

The sample-efficiency improvement gained via VPACE allowed us to test it on real-life tasks (two right columns of \cref{fig:env_examples_main_paper}), where the bottom right plots in \cref{fig:perf_results_all_summary} demonstrate that VPACE can solve real-world tasks in only a few hours of real execution time, whereas SQIL does not solve either task.
Our real robot tasks execute at 5 Hz, and, with a small amount of time to allow for environment resetting, 1000 environment steps corresponds to roughly 4 real minutes, meaning \texttt{RealDrawer} is learned by VPACE in about 100 minutes, while \texttt{RealDoor} is learned in about 200 minutes.
The real world examples of success, for both main tasks and auxiliary tasks, are collected in less than a minute.
Our open-source code contains further implementation details for our real world setup.

\subsection{Analysis and Additional Results}

\subsubsection{Q-Value Overestimation and Value Penalization}
\label{sec:val_pen_and_cal_of_qvalues}

\begin{figure}
    \centering
    \includegraphics[width=.8\linewidth]{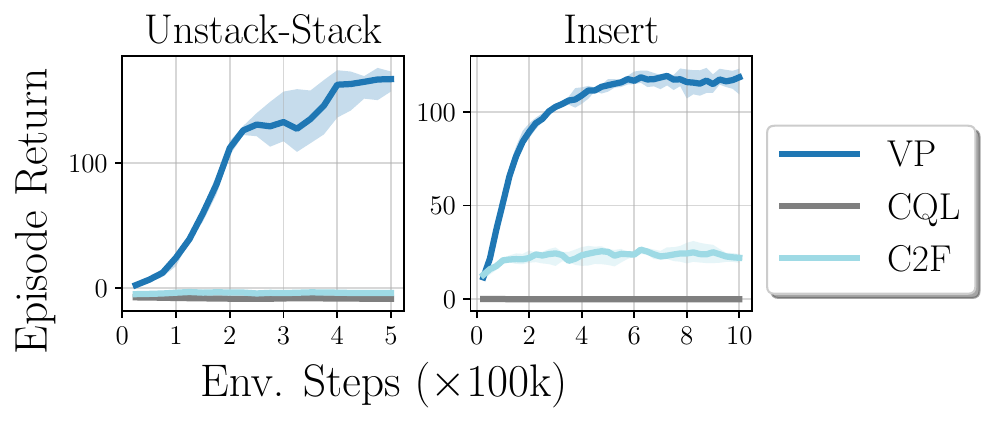}
    \caption{
        Results for different example-based approaches to regularizing Q estimates.
        We measure performance using return instead of success rate as the alternatives achieved no success at all.
    }
    \label{fig:perf_results_abl_reg}
    \vspace{-12pt}
\end{figure}

\begin{figure}
    \centering
    \includegraphics[width=.95\linewidth]{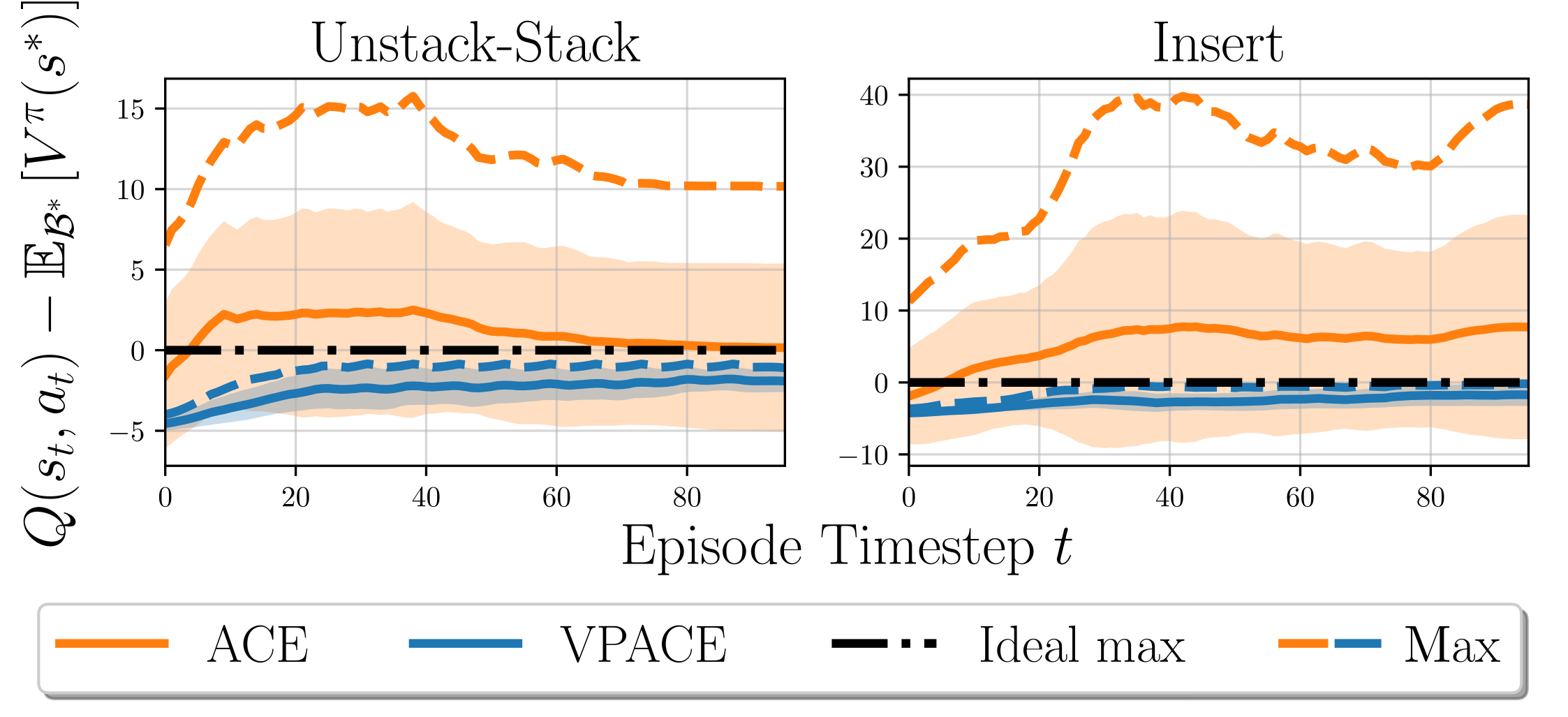}
    \caption{
        Difference between Q-values and expected value for example states for partially trained agents (snapshot from 300k environment steps) for a single episode rollout.
        The dashed lines show the maximum output across the seeds, and the shaded regions show the standard deviation between seeds.
        VP ensures that the maximum stays below 0, verifying that $y(s, s') \leq y(s^*, s^*)$.
    }
    \label{fig:per_ts_q_values_main}
    \vspace{-6pt}
\end{figure}

\begin{figure}
    \centering
    \includegraphics[width=.95\linewidth]{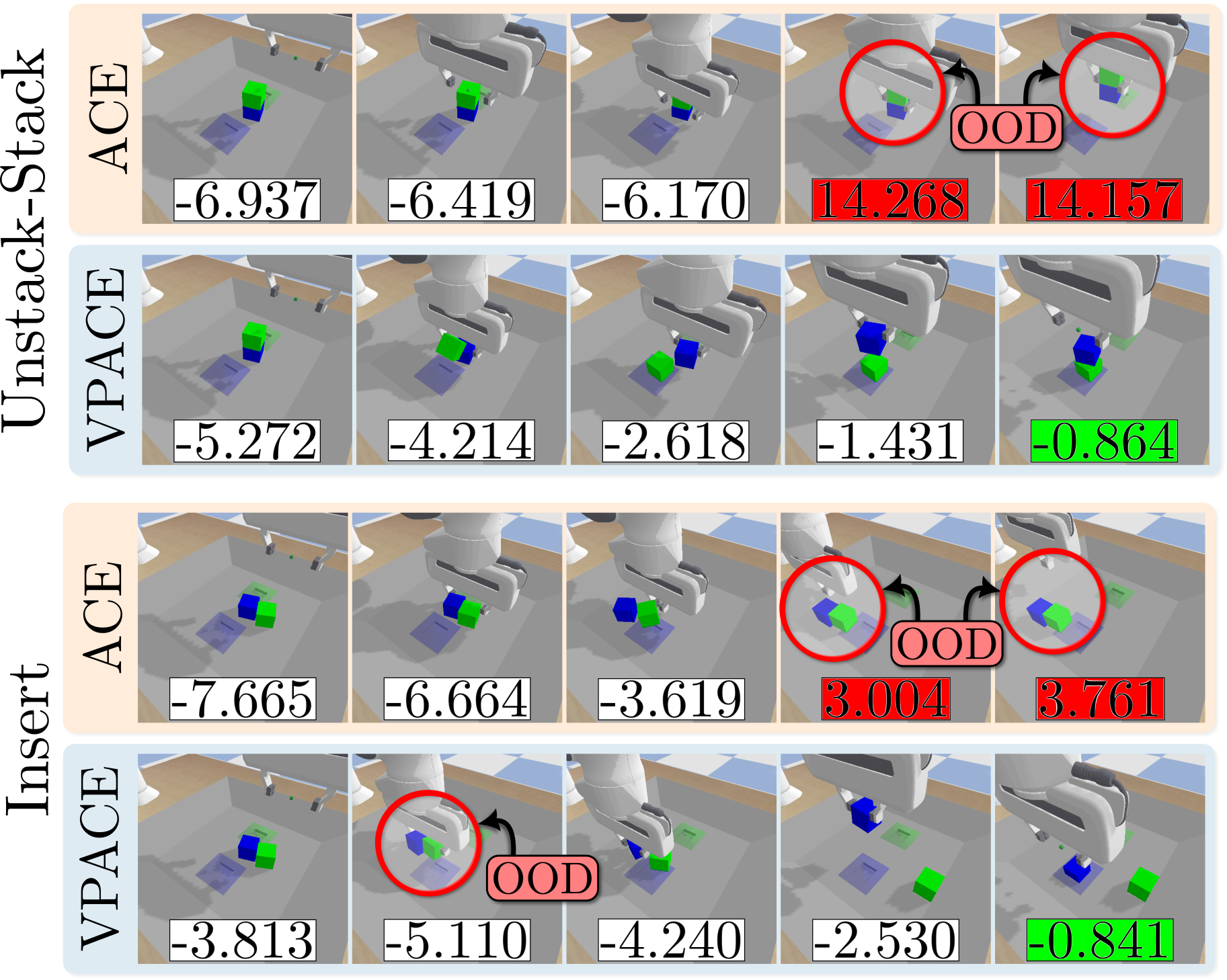}
    \caption{
        $Q(s_t, a_t) - \ExpectationSamp{\buffer^*}{V^{\pi} (s^*)}$ for parts of episodes used to generate the results from \cref{fig:per_ts_q_values_main}.
        It appears that the use of ACE without VP results in overestimated Q-values for out-of-distribution (OOD) $(s_t,a_t)$ pairs (e.g., grasping both blocks simultaneously, or nearly inserting the incorrect block), resulting in highly suboptimal policies.
        VPACE, on the other hand, correctly learns to label a potentially OOD state with low value.
    }
    \label{fig:per_ts_q_ood_example}
    \vspace{-12pt}
\end{figure}

To verify that VP enforces $y(s, s') \leq y(s^*, s^*)$, we took snapshots of each learned \texttt{Unstack-Stack} and \texttt{Insert} agent, for both VPACE and ACE, at 300k steps and ran each policy for a single episode, recording per-timestep Q-values.
Instead of showing Q-values directly, in \cref{fig:per_ts_q_values_main,fig:per_ts_q_ood_example}, we show $Q(s_t, a_t) - \mathbb{E}_{s^* \sim \buffer^*} \left[ V(s^*) \right]$, which should be at most $0$ for $y(s, s') \leq y(s^*, s^*)$ to hold.
ACE clearly violates  $y(s, s') \leq y(s^*, s^*)$, while VPACE does not.
Furthermore, \cref{fig:per_ts_q_ood_example} shows examples of the consequences of overestimated Q-values: for $(s_t,a_t)$ pairs that appear to be out-of-distribution (OOD), $Q(s_t, a_t) -\mathbb{E}_{s^* \sim \buffer^*} \left[ V(s^*) \right] > 0$, and the resulting policy reaches these states instead of the true goal.

To understand the importance of the $y(s, s') \leq y(s^*, s^*)$ constraint, we investigate other choices of regularization techniques (RQ3), in particular a L2 regularizer on Q-values (\textbf{C2F}, \cite{jamesCoarsetoFineQAttentionEfficient2022}) and a regularizer that penalizes the Q-values of out-of-distribution actions (\textbf{CQL}, \cite{kumarConservativeQLearningOffline2020}).
\cref{fig:perf_results_abl_reg} indicates that using other existing regularization techniques does not enable ACE to perform well.
We suspect that the L2 regularizer may be too harsh of a penalty on all learning, while \cite{kumarConservativeQLearningOffline2020} does not have any penalization on the magnitudes of Q-values in general, meaning that they can potentially still have uncontrollable bootstrapping error.

\subsubsection{Comparison to Full Demonstrations and True Rewards}
\label{subsec:comparison_to_full_demos_and_true_rews}

A practitioner may ask how our method compares to using traditional forms of feedback, such as using full-expert trajectories and inverse reinforcement learning (IRL), or using RL with true sparse rewards (RQ4).
To test the former, we include two variants: a buffer containing both 200 examples and 200 $(s,a)$ pairs from expert trajectories of states only (\textbf{+Trajs}), and the same for both states and actions (\textbf{+Trajs \& Acts}), resulting in two variants of LfGP \cite{ablettLearningGuidedPlay2023} with VP.
For sparse rewards, we use the provided sparse-reward function from the task (\textbf{SAC-X}), which is scheduled auxiliary control \cite{riedmillerLearningPlayingSolving2018} with VP, although VP is based on the maximum theoretical return in these experiments, rather than $y(s, s') \leq y(s^*, s^*)$ since $s^*$ are not available to SAC-X.

\begin{figure}
	\centering
	\includegraphics[width=.9\linewidth]{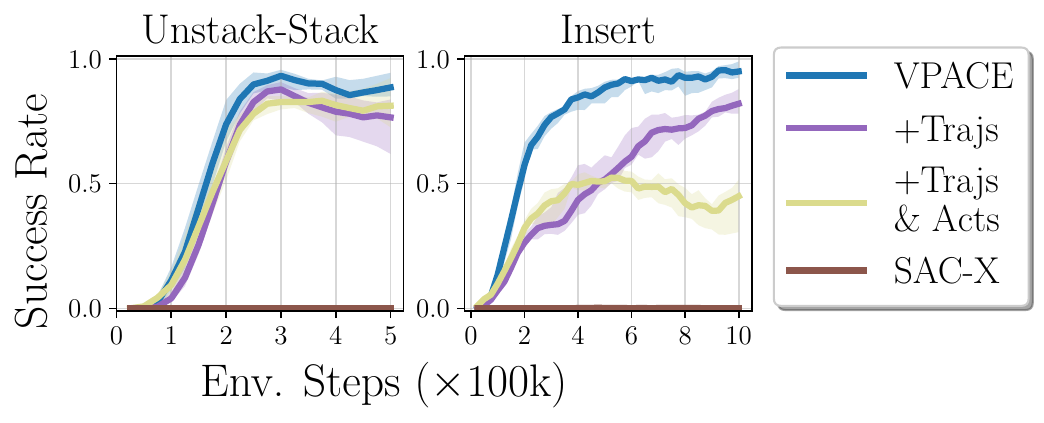}
	\caption{
		Results for changes in the form of feedback.
		SAC-X indicates the use of true environment sparse rewards only.
	}
	\label{fig:perf_results_abl_expert}
	\vspace{-12pt}
\end{figure}

\cref{fig:perf_results_abl_expert} shows that the peak performance is \textit{reduced} when learning with expert trajectories.
We hypothesize that the divergence minimization objective leads to an effect, commonly seen with dense reward functions, known as reward hacking \cite{skalseDefiningCharacterizingReward2022}.
This effect was also previously shown to occur in inverse RL \cite{ablettLearningGuidedPlay2023}.
In short, the agent receives the same positive reward for reaching intermediate states in the demonstration as it does for reaching final states, despite the fact that intermediate states may not lead to the optimal final states.
This result suggests that inverse RL/adversarial IL can be significantly improved by switching to example-based control.
Learning with a sparse-reward function only does not accomplish the main task at all in either \texttt{Unstack-Stack} nor \texttt{Insert}, exhibiting substantially poorer performance than all other baselines.
This result matches previous research in which it has been shown that a demonstration buffer can significantly improve sample efficiency in RL \cite{nairOvercomingExplorationReinforcement2018,kalashnikovQTOptScalableDeep2018}.
Our results show that an example buffer may provide a similar effect, and may even be a more optimal form of guidance than full demonstrations.

\begin{figure}
	\centering
	\includegraphics[width=.9\linewidth]{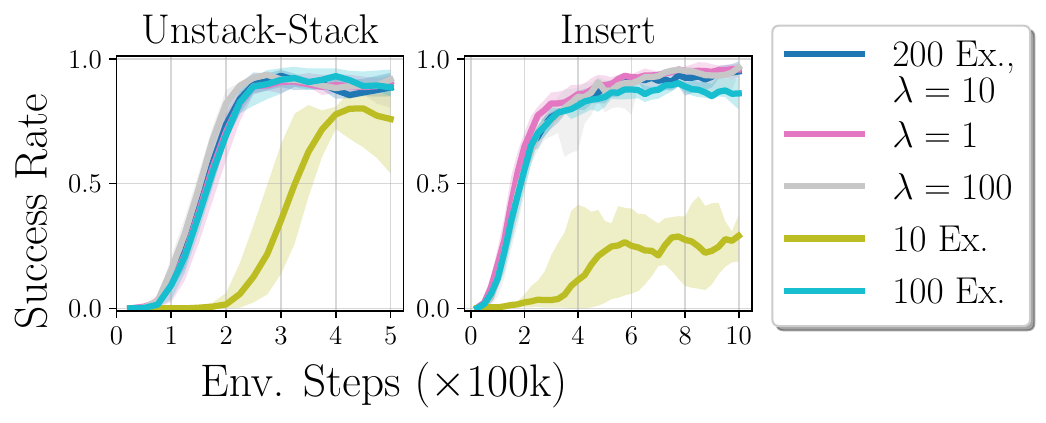}
	\caption{
		Variations of success example quantity and $\lambda$ value.
	}
	\label{fig:abl_dquant_lambda}
	\vspace{-12pt}
\end{figure}

\subsubsection[Expert Data Quantity and Lambda Value]{Expert Data Quantity and $\lambda$ Value}
\label{subsec:expert_data_quantity_and_lambda_value}

Finally, to examine the robustness of our approach, we also compare various values for the quantity of examples of success (200 / task for main experiments) as well as the value of $\lambda$, which controls the strength of value penalization (10.0 for main experiments).
\cref{fig:abl_dquant_lambda} shows that changing the value of $\lambda$ to 1 and 100 has no effect on performance, likely because it has no effect if the value constraint is not violated.
Dropping from 200 to 100 examples has only a minor negative impact on performance in \texttt{Insert}, while the drop to 10 examples has a significant negative effect.
However, even with only 10 examples, performance on \texttt{Unstack-Stack} still far exceeds all other baselines.

\section{Conclusion}
\label{sec:conclusion}

In this work, we presented VPACE---value-penalized auxiliary control from examples, where we coupled scheduled auxiliary control with value penalization in the example-based setting to significantly improve learning efficiency.
We showed that VPACE resolves Q overestimation, greatly improves the sample efficiency of example-based control against a wide set of baselines, and shows improved performance compared with with other forms of feedback.
Opportunities for future work include the further investigation of learned approaches to scheduling, as well as autonomously generating auxiliary task definitions.

\section*{Acknowledgements}
\label{sec:acknowledgements}

We gratefully acknowledge the Digital Research Alliance of Canada and NVIDIA Inc., who provided the GPUs used in this work through their Resources for Research Groups Program and their Hardware Grant Program, respectively.
This work was supported in part by the Natural Sciences and Engineering Research Council of Canada (NSERC).
Jonathan Kelly gratefully acknowledges support from the Canada Research Chairs program.

\bibliographystyle{IEEEtran}
\bibliography{IEEEabrv, vpace}  %

\newpage
\appendices
\section{Reward Model Formulations}
\label{app:reward_model_options}
In this section, we investigate approaches to off-policy reinforcement learning (RL) studied in this work, modified to accommodate an unknown $R$ and the existence of an example state buffer $\buffer^*$.

\subsection{Learning a Reward Function}
\label{sec:learning_a_reward_function}
The most popular approach to reward modelling, known as inverse RL, tackles an unknown $R$ by explicitly learning a reward model.
Modern approaches under the class of adversarial imitation learning (AIL) algorithm aim to learn both the reward function and the policy simultaneously.
In AIL, the learned reward function, also known as the discriminator, aims to differentiate the occupancy measure between the state-action distributions induced by expert and the learner.
The learned reward function $\hat{R}$ is derived from the minimax objective \cite{kostrikovDiscriminatorActorCriticAddressingSample2019,hoGenerativeAdversarialImitation2016,fuLearningRobustRewards2018}:
\begin{equation}
\label{eq:example-based-ail}
\begin{aligned}
    \mathcal{L}(D) = \mathbb{E}_{\buffer}\left[ \log \left( 1 - D(s)\right) \right] + \mathbb{E}_{\buffer^*}\left[ \log \left( D(s^*)\right) \right],
\end{aligned}
\end{equation}
where $D$ attempts to differentiate the occupancy measure between the state distributions induced by $\buffer^*$ and $\buffer$.
The output of $D(s)$ is used to define $\hat{R}(s)$, which is then used for updating the Q-function.

In example-based control, the discriminator $D$ provides a smoothed label of success for states, thus its corresponding reward function can provide more density than a typical sparse reward function, making this approach an appealing choice.
Unfortunately, a learned discriminator can suffer from the deceptive reward problem, as previously identified in \cite{ablettLearningGuidedPlay2023}, and this problem is exacerbated in the example-based setting.
In the following sections, we describe options to remove the reliance on separately learned discriminators.

\subsection{Discriminator-Free Reward Labels with Mean Squared Error TD Updates}
\label{sec:reward_labels_regression}
A simple alternative to using a discriminator as a reward model was initially introduced as soft-Q imitation learning (SQIL) in \cite{reddySQILImitationLearning2020}.
In standard AIL algorithms, $D$ is trained separately from $\pi$ and $Q^\pi$, where $D$ is trained using data from both $\buffer$ and $\buffer^*$, whereas $\pi$ and $Q^\pi$ are trained using data exclusively from $\buffer$.
However, most off-policy algorithms do not \textit{require} this choice, and approaches such as \cite{ablettLearningGuidedPlay2023} and \citeapp{vecerikLeveragingDemonstrationsDeep2018} train $Q$, and possibly $\pi$, using data from both $\buffer$ and $\buffer^*$.
It is unclear why this choice is often avoided in AIL, but it might be because it can introduce instability due to large discrepancy in magnitudes for Q targets given data from $\buffer$ and $\buffer^*$.
Sampling from both buffers, we can define $\hat{R}(s_t, a_t)$ to be labels corresponding to whether the data is sampled from $\buffer$ or $\buffer^*$---in SQIL, the labels are respectively 0 and 1.
The full training objective resembles the minimax objective in \cref{eq:example-based-ail} where we set $D(s^*) = 1$ and $D(s) = 0$.

The resulting reward function is the expected label $\hat{R}(s, a) = \mathbb{E}_{ \hat{\buffer} \sim \text{Categorical}_s(\{\buffer, \buffer^*\})}\left[ \boldsymbol{1}(\hat{\buffer} = \buffer^*)\right]$,
where $\text{Categorical}_s$ is a categorical distribution corresponding to the probability that $s$ belongs to buffers $\buffer, \buffer^*$.
Consequently, only successful states $s^*$ yields positive reward and the corresponding optimal policy will aim to reach a successful state as soon as possible.
If we further assume that $s^*$ transitions to itself, then for policy evaluation with mean-squared error (MSE), we can write the temporal difference (TD) target, $y$, of Q-updates with:

\begin{tabularx}{\textwidth}{@{}XX@{}}
  \begin{align}
    \label{eq:sqil_policy_bellman}
     y(s, s') = \gamma \mathbb{E}_{a'}\left[ Q(s', a') \right],
  \end{align} &
  \begin{align}
    \label{eq:sqil_expert_bellman}
    y(s^*, s^*) = 1 +  \gamma \mathbb{E}_{a'}\left[ Q(s^*, a') \right],
  \end{align}
\end{tabularx}
where $(s, \cdot, s') \sim \buffer$, $a' \sim \pi(a' \vert s')$, and $s^* \sim \buffer^*$.

This approach reduces complexity as we no longer explicitly train a reward model, and also guarantees discrimination between data from $\buffer$ and $\buffer^*$.

\subsection{Discriminator-Free Reward Labels with Binary Cross Entropy TD Updates}
\label{sec:reward_labels_classification}

\cite{eysenbachReplacingRewardsExamples2021} introduced recursive classification of examples (RCE), a method for learning from examples of success.
RCE mostly follows the approach outlined above in \cref{sec:reward_labels_regression} but uses a weighted binary cross-entropy (BCE) loss with weights $1 + \gamma w$ for data from $\buffer$ and $1 - \gamma$ for data from $\buffer^*$.
The TD targets are also changed from \cref{eq:sqil_policy_bellman} and \cref{eq:sqil_expert_bellman} to

\begin{tabularx}{\textwidth}{@{}XX@{}}
  \begin{align}
    \label{eq:rce_policy_bellman}
     y(s, s') = \gamma w(s') / (1 + \gamma w(s')),
  \end{align} &
  \begin{align}
    \label{eq:rce_expert_bellman}
    y(s^*, s^*) = 1,
  \end{align}
\end{tabularx}
where $w(s') = V^{\pi}(s')/(1 - V^{\pi}(s'))$.

This approach can also be made closer to SQIL by removing the change in weights and by leaving \cref{eq:rce_policy_bellman} as \cref{eq:sqil_policy_bellman}.
This makes it equivalent to SQIL, apart from removing bootstrapping from \cref{eq:sqil_expert_bellman}, and using a BCE instead of MSE for TD updates.

A major benefit of this approach, compared with the MSE TD updates in \cref{sec:reward_labels_regression}, is that $y(s, s') \leq y(s^*, s^*)$ is always enforced at every update, meaning that our approach to value penalization would provide no extra benefit.
Nonetheless, our results from \cref{sec:sqil_vs_rce} show that SQIL, even \textit{without} value-penalization, almost always outperforms both RCE and SQIL with BCE loss (referred to SQIL-BCE here).

\section{Additional Environment, Algorithm, and Implementation Details}
\label{sec:add_env_alg_impl_details}

The following sections contain further details of the environments, tasks, auxiliary tasks, algorithms, and implementations used in our experiments.

\subsection{Additional Environment Details}

Compared with the original Panda tasks from LfGP \cite{ablettLearningGuidedPlay2023}, we switch from 20Hz to 5Hz control (finding major improvements in performance for doing so), improve handling of rotation symmetries in the observations, and remove the force-torque sensor since it turned out to have errors at low magnitudes.
Crucially, these modifications did not require training new expert policies, since the same final observation states from the full trajectory expert data from \cite{ablettLearningGuidedPlay2023} remained valid.
Compared with the original LfGP tasks, we also remove \texttt{Move-block} as an auxiliary task from \texttt{Stack}, \texttt{Unstack-Stack}, \texttt{Bring} and \texttt{Insert}, since we found a slight performance improvement for doing so, and add \texttt{Reach}, \texttt{Lift}, and \texttt{Move-block} as main tasks.
The environment was otherwise identical to how it was implemented in LfGP, including allowing randomization of the block and end-effector positions anywhere above the tray, using delta-position actions, and using end-effector pose, end-effector velocity, object pose, object velocity, and relative positions in the observations.
For even further details of the original environment, see \cite{ablettLearningGuidedPlay2023}.

Since the Sawyer tasks from \cite{eysenbachReplacingRewardsExamples2021, yuMetaWorldBenchmarkEvaluation2019} only contain end-effector position and object position by default, they do not follow the Markov property.
To mitigate this, we train all algorithms in the Sawyer tasks with frame-stacking of 3 and add in gripper position to the observations, since we found that this, at best, improved performance for all algorithms, and at worst, kept performance the same.
We validate that this is true by also performing experiments using the original code and environments from \cite{eysenbachReplacingRewardsExamples2021}, unmodified, where the results of RCE without these modifications are presented in \cref{fig:rce_hand_theirs_performance}, with results comparable to or poorer than our own RCE results.

\subsection{Delta-Position Adroit Hand Environments}
\label{sec:dp_adroit_env_details}

\begin{figure}
    \centering
    \includegraphics[width=\linewidth]{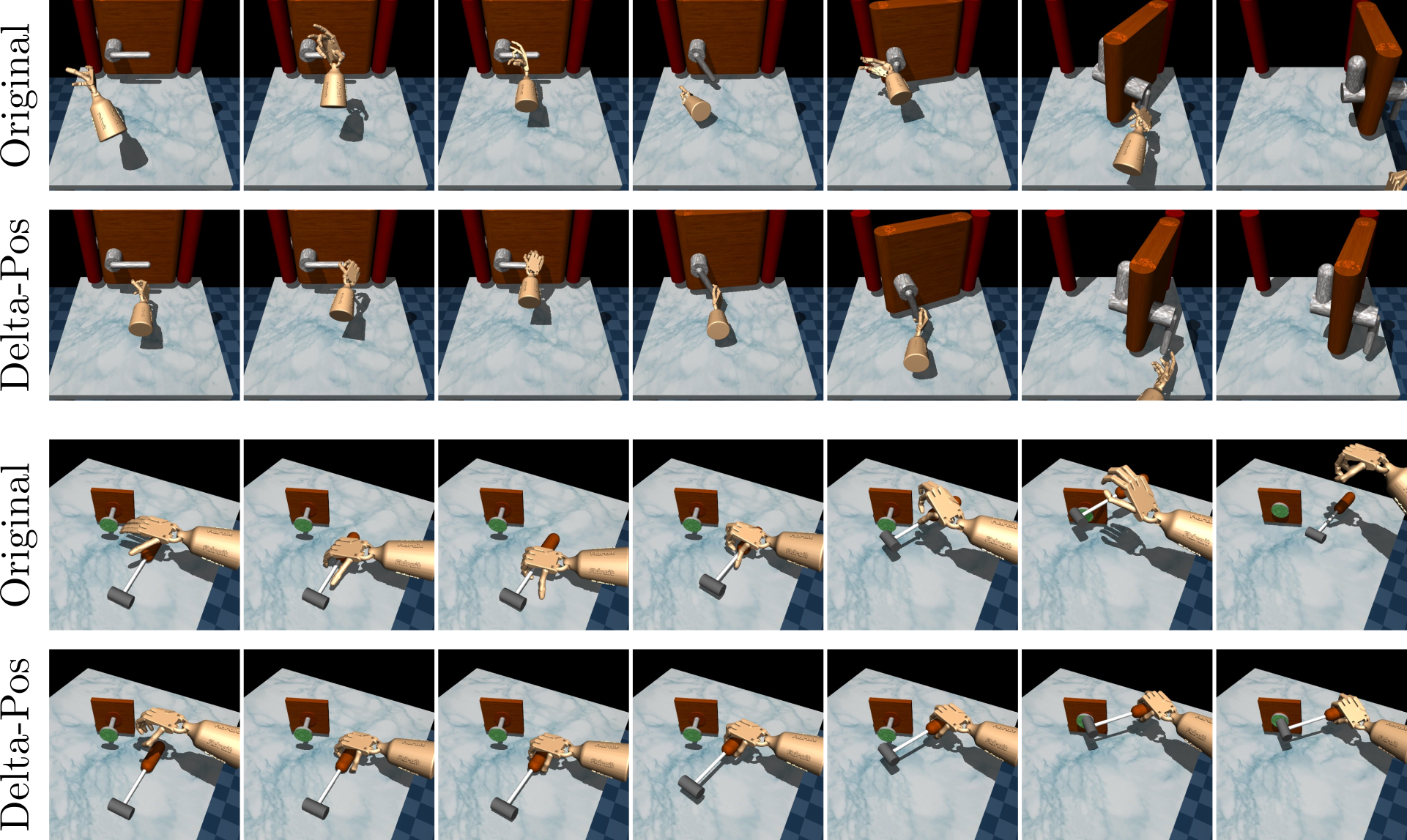}
    \caption{
        Learned VPACE-SQIL policies at the final training step for, from top to bottom, \texttt{door-human-v0}, \texttt{door-human-v0-dp}, \texttt{hammer-human-v0}, and \texttt{hammer-human-v0-dp}.
        Although the original versions of the environments are solved, the absolute position action space allows policies to execute very coarse actions that exploit the simulator (above, hitting the handle without grasping it and throwing the hammer, respectively), and would almost certainly cause damage to a real environment.
    }
    \label{fig:hand_dp_traj_examples}
\end{figure}

The Adroit hand tasks from \cite{rajeswaran*LearningComplexDexterous2018} use absolute positions for actions.
This choice allows even very coarse policies, with actions that would be unlikely to be successful in the real world, to learn to complete \texttt{door-human-v0} and \texttt{hammer-human-v0}, and also makes the intricate exploration required to solve \texttt{relocate-human-v0} very difficult.
Specifically, VPACE-SQIL and several other baselines achieve high return in \texttt{door-human-v0} and \texttt{hammer-human-v0}, but the learned policies use unrealistic actions that exploit simulator bugs.
As well, no methods are able to achieve any return in \texttt{relocate-human-v0} in 1.5M environment steps.

In the interest of solving \texttt{relocate-human-v0} and learning more skillful policies, we generated modified versions of these environments with delta-position action spaces.
Furthermore, in \texttt{relocate-human-v0-najp-dp}, the action space rotation frame was changed to be in the palm, rather than at the elbow, and, since relative positions between the palm, ball, and relocate goal are included as part of the state, we removed the joint positions from the state.
In our experiments, these modified environments were called \texttt{door-human-v0-dp}, \texttt{hammer-human-v0-dp}, and \texttt{relocate-human-v0-najp-dp}.
See \cref{fig:hand_dp_traj_examples} for a comparison of learned policies in each version of these environments.

\subsection{Real World Environment Details}

\begin{figure}
    \centering
    \includegraphics[width=\linewidth]{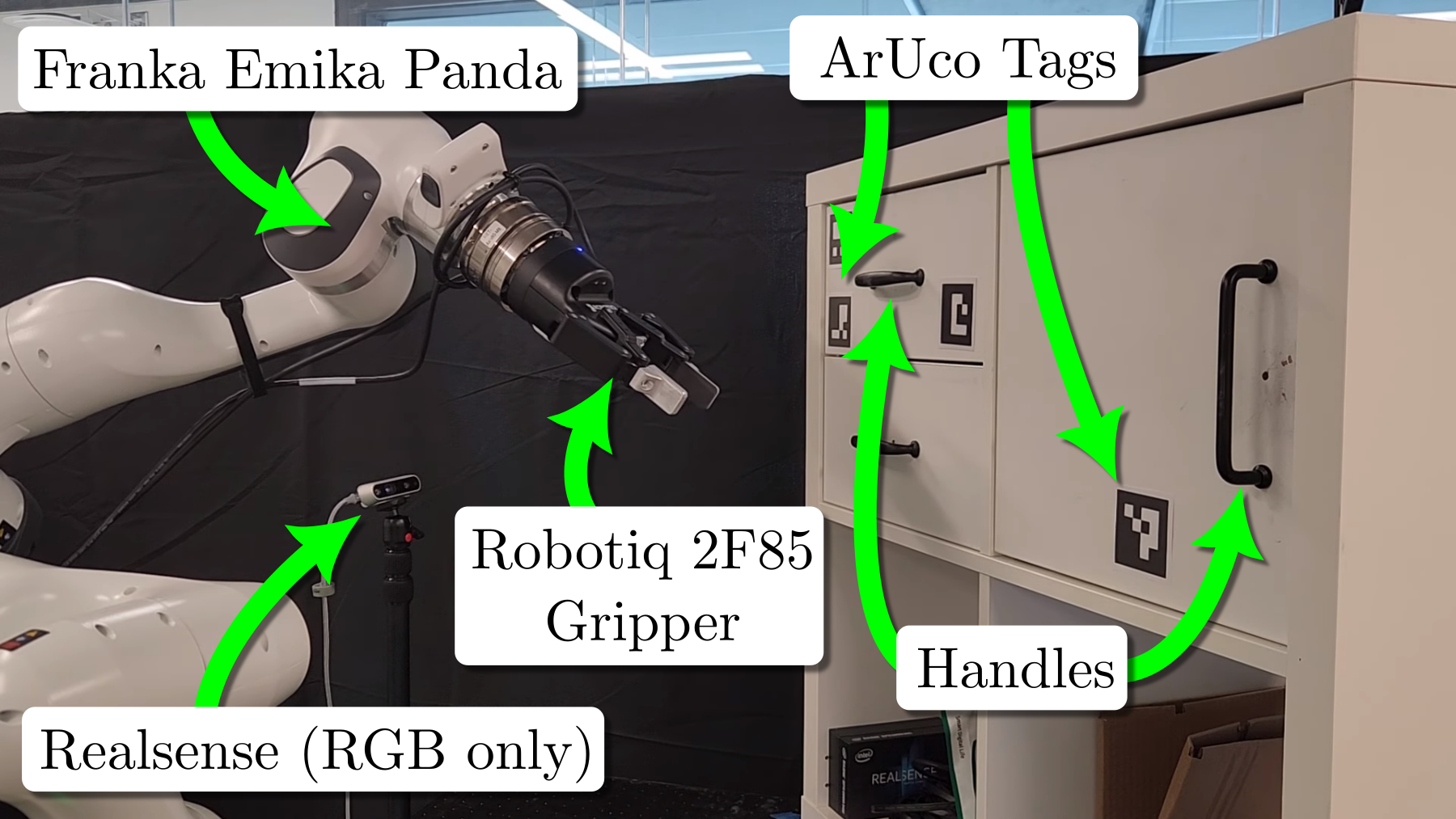}
    \caption{
        Experimental setup for our real environment and our \texttt{RealDrawer} and \texttt{RealDoor} tasks.
        The robot has a force-torque sensor attached, but it is not used for our experiments.
    }
    \label{fig:real_setup}
\end{figure}

\cref{fig:real_setup} shows our experimental platform and setup for our two real world tasks.
In both \texttt{RealDrawer} and \texttt{RealDoor}, the observation space contains the end-effector position, an ArUco tag \cite{garrido-juradoAutomaticGenerationDetection2014} to provide drawer or door position (in the frame of the RGB camera; we do not perform extrinsic calibration between the robot and the camera), and the gripper finger position.
Observations also include two stacked frames, in lieu of including velocity, to better follow the Markov property.
The action space in both contains delta-positions and a binary gripper command for opening and closing.
The action space for \texttt{RealDrawer} is one-dimensional (allowing motion in a line), while the action space for \texttt{RealDoor} is two-dimensional (allowing motion in a plane).
The initial state distribution for \texttt{RealDrawer} allows for initializing the end-effector anywhere within a 10 cm line approximately 25 cm away from the drawer handle when closed.
For \texttt{RealDoor}, the initial state distribution is a 20 cm $\times$ 20 cm square, approximately 20cm away from the door handle when closed.
Actions are supplied at 5 Hz.

For both environments, for evaluation only, success is determined by whether the drawer or door is fully opened, as detected by the absolute position of the ArUco tag in the frame of the RGB camera.
Our robot environment code is built on Polymetis \citeapp{Polymetis2021}, and uses the default hybrid impedance controller that comes with the library.
To reduce environmental damage from excessive forces and torques, we reduced Cartesian translational stiffness in all dimensions from 750 N/m to 250 N/m, and the force and torque limits in all dimensions from 40 N and 40 Nm to 20 N and 20 Nm.

\subsection{Additional Task Details}
\label{sec:additional_task_details}

Success examples for the Panda environments were gathered by taking $s_T$ from the existing datasets provided by \cite{ablettLearningGuidedPlay2023}.
Success examples for main tasks from the Sawyer environments were generated using the same code from \cite{eysenbachReplacingRewardsExamples2021}, in which the success examples were generated manually given knowledge of the task.
Auxiliary task data was generated with a similar approach.
Success examples for the Adroit hand environments were generated from the original human datasets provided by \cite{rajeswaran*LearningComplexDexterous2018}.
Success examples for our real world tasks were generated by manually moving the robot to a small set of successful positions for each auxiliary task and main task.
All Panda main tasks use the the auxiliary tasks \textit{release}, \textit{reach}, \textit{grasp}, and \textit{lift}.

There are two specific nuances that were left out of the main text for clarity and brevity:
(i) the \texttt{Reach} main task only uses \textit{release} as an auxiliary task (since it also acts as a ``coarse" reach), and
(ii) half of the \textit{release} dataset for each task is specific to that task (e.g., containing insert or stack data), as was the case in the original datasets from \cite{ablettLearningGuidedPlay2023}.
For the Sawyer, Hand, and real Panda environments, because the observation spaces are not shared, each task has its own separate \textit{reach} and \textit{grasp} data.

\subsection{Additional Algorithm Details}
\label{sec:additional_alg_details}

\begin{algorithm}[!ht]
\caption{Value-Penalized Auxiliary Control from Examples (VPACE)}
\label{alg:vpace}
\textbf{Input}: Example state buffers $\buffer^*_{\text{main}}, \buffer^*_{1}, \dots, \buffer^*_{K}$, scheduler period $\xi$, sample batch size $N$ and $N^*$, and discount factor $\gamma$\\
\textbf{Parameters}: Intentions $\pi_\tasks$ with corresponding Q-functions $Q_\tasks$ \textcolor{blue}{(and optionally discriminators $D_\tasks$)}, and scheduler $\pi_S$ (e.g. with Q-table $Q_S$)
\begin{algorithmic}[1] %
\STATE Initialize replay buffer $\mathcal{B}$ \\
\FOR{$t = 1, \dots,$}
    \STATE \# Interact with environment
    \STATE For every $\xi$ steps, select intention $\pi_\tasks$ using $\pi_S$
    \STATE Select action $a_t$ using $\pi_\tasks$
    \STATE Execute action $a_t$ and observe next state $s'_t$
    \STATE Store transition $\langle s_t, a_t, s'_t \rangle$ in $\mathcal{B}$
    \STATE
    \textcolor{blue}{
    \STATE \# Optionally update discriminator $D_{\tasks'}$ for each task $\tasks'$
    \STATE Sample $\left\{ s_i \right\}_{i=1}^{N} \sim \mathcal{B}$
    \FOR{each task $\tasks'$}
        \STATE Sample $\left\{ s^*_i \right\}_{i=1}^{N^*} \sim \buffer^*_k$
        \STATE Update $D_{\tasks'}$ following \cref{eq:example-based-ail} using GAN + Gradient Penalty
    \ENDFOR}
    \STATE
    \STATE \# Update intentions $\pi_{\tasks'}$ and Q-functions $Q_{\tasks'}$ for each task $\tasks'$
    \STATE Sample $\left\{ (s_i, a_i) \right\}_{i=1}^{N} \sim \mathcal{B}$
    \FOR{each task $\tasks'$}
        \STATE Sample $\left\{ s^*_i \right\}_{i=1}^{N^*} \sim \mathcal{B^*_{\tasks'}}$
        \STATE Sample $a^*_i \sim \pi_{\tasks'}(s^*_i)$ for $i = 1, \dots, N^*$

        \STATE Compute rewards $\hat{R}_{\tasks'}(s_i)$ and $\hat{R}_{\tasks'}(s^*_j)$ for $i = 1, \dots, N$ and $j = 1, \dots, N^*$
        \STATE \# Compute value penalization terms, see \cref{sec:maintaining_qmax_qmin_estimates}
        \STATE Compute $Q^{\pi_{\tasks'}}_\text{max} = \frac{1}{N^*} \sum_{j=1}^{N^*} Q(s^*_j, a^*_j)$
        \STATE Compute $Q^{\pi_{\tasks'}}_\text{min} = \min(\wedge_{i=1}^N \hat{R}_{\tasks'}(s_i), \wedge_{j=1}^{N^*} \hat{R}_{\tasks'}(s^*_j)) / (1 - \gamma)$
    \ENDFOR
    \STATE Update $\pi$ following \cref{eq:multi_pol_obj_app}
    \STATE Update $Q$ following \cref{eq:multi_q_obj_app} with value penalization \cref{eq:value_penalized_q_update_app}
    \STATE
    \STATE \# \textit{Optional }Update learned scheduler $\pi_S$
    \IF{at the end of effective horizon}
        \STATE Compute main task return $G_{\tasks_\text{main}}$ using reward estimate from $D_\text{main}$
        \STATE Update $\pi_S$ (e.g. update Q-table $Q_S$ using EMA and recompute Boltzmann distribution)
    \ENDIF
\ENDFOR
\end{algorithmic}
\vspace{2mm}
\end{algorithm}

\begin{table}
    \centering
    \small
    \begin{tabularx}{\linewidth}{l|YYYYY}  %
        Algorithm & Value Penalization & Sched. Aux. Tasks & Reward Model & TD Error Loss & Source
        \\\midrule
        VPACE & \cmark & \cmark & SQIL & MSE & Ours \\
        ACE & \xmark & \cmark & SQIL & MSE & Ours \\
        VPACE-DAC & \cmark & \cmark & DAC & MSE & Ours \\
        ACE-RCE & \xmark & \cmark & RCE & BCE & Ours \\
        VP-SQIL & \cmark & \xmark & SQIL & MSE & Ours \\
        DAC & \xmark & \xmark & DAC & MSE & Ours \\
        SQIL & \xmark & \xmark & SQIL & MSE & Ours \\
        RCE & \xmark & \xmark & RCE & BCE & Ours \\
        SQIL+RND & \xmark & \xmark & SQIL & MSE & Ours \\
        RCE (theirs) & \xmark & \xmark & RCE & BCE & \cite{eysenbachReplacingRewardsExamples2021} \\
        SQIL-BCE & \xmark & \xmark & SQIL & BCE & \cite{eysenbachReplacingRewardsExamples2021} \\
    \end{tabularx}
    \caption{
        Major differences between algorithms studied in this work.
        MSE refers to mean squared error, while BCE refers to binary cross entropy.
    }
    \label{tab:algorithm_table}
\end{table}

\cref{alg:vpace} shows a summary of VPACE, built on LfGP \cite{ablettLearningGuidedPlay2023} and SAC-X \cite{riedmillerLearningPlayingSolving2018}.
As a reminder, our multi-policy objective function is
\begin{equation}
    \label{eq:multi_pol_obj_app}
    \mathcal{L}(\pi; \tasks) = \mathbb{E}_{\buffer, \pi_\tasks}\left[ Q_\tasks(s, a) \right],
\end{equation}
our multi-q objective function is
\begin{align}
\begin{split}
    \label{eq:multi_q_obj_app}
    \mathcal{L}(Q; \tasks) = &\mathbb{E}_{\buffer, \pi_\tasks}\left[ (Q_\tasks(s, a) - y_\tasks(s, s'))^2 \right]\\
    &+ \mathbb{E}_{\buffer^*_\tasks, \pi_\tasks}\left[ (Q_\tasks(s^*, a) - y_\tasks(s^*, s^*))^2 \right],
\end{split}
\end{align}
and our value-penalized update is
\begin{align}
\begin{split}
\label{eq:value_penalized_q_update_app}
    \mathcal{L}^\pi_{\text{pen}}(Q) = \lambda \mathbb{E}_{\buffer} [ &\left( \max(Q(s,a) - Q^\pi_{\text{max}}, 0) \right)^2\\
    &+ \left( \max(Q^\pi_{\text{min}} - Q(s,a), 0) \right)^2 ],
\end{split}
\end{align}
where 
\begin{equation}
\label{eq:penalty_q_min}
     Q^\pi_{\text{min}} = \hat{R}_{\text{min}} / (1 - \gamma),
\end{equation}
and
\begin{equation}
\label{eq:penalty_q_max}
    Q^\pi_{\text{max}} = \mathbb{E}_{\buffer^*} \left[ V^\pi(s^*) \right].
\end{equation}

\cref{tab:algorithm_table} shows a breakdown of some of the major differences between all of the algorithms studied in this work.

\subsection{Additional Implementation Details}
\label{sec:additional_impl_details}

In this section, we list some specific implementation details of our algorithms.
We only list parameters or choices that may be considered unique to this work, but a full list of all parameter choices can be found in our code.
We also provide the VPACE pseudocode in \cref{alg:vpace}, with \textcolor{blue}{blue text} only applying to learned discriminator-based reward functions (see \cref{app:reward_model_options} for various reward models).

\begin{table}
    \centering
    \small
    \caption{
        Hyperparameters shared between all algorithms, unless otherwise noted.
    }
    \begin{tabularx}{\linewidth}{XY}  %
        \toprule
            \textit{General} & \\
            Total Interactions & Task-specific (see \cref{sec:add_performance_results}) \\
            Buffer Size & Same as total interactions \\
		Buffer Warmup & 5k \\
		Initial Exploration & 10k \\
            Evaluations per task & 50 (Panda), 30 (Sawyer/Adroit) \\
            Evaluation frequency & 25k (Panda), 10k (Sawyer/Adroit) \\
        \midrule
        \midrule
		\textit{Learning} &  \\
		$\gamma$ & 0.99  \\
		$\buffer$ Batch Size & 128  \\
		$\buffer^*$ Batch Size & 128  \\
		$Q$ Update Freq. & 1 (sim), 4 (real)  \\
		Target $Q$ Update Freq. & 1 (sim), 4 (real) \\
		$\pi$ Update Freq. & 1 (sim), 4 (real) \\
		Polyak Averaging & 1e-3 \\
            $Q$ Learning Rate & 3e-4 \\
		$\pi$ Learning Rate & 3e-4 \\
            $D$ Learning Rate & 3e-4 \\
            $\alpha$ Learning Rate & 3e-4 \\
            Initial $\alpha$ & 1e-2 \\
            Target Entropy & $-\text{dim}(a)$ \\
            Max. Gradient Norm & 10 \\
            Weight Decay ($\pi,Q,D$) & 1e-2 \\
            $D$ Gradient Penalty & 10 \\
            Reward Scaling & 0.1 \\
            SQIL Labels (\cref{sec:additional_impl_details} & $(-1, 1)$ \\
            Expert Aug. Factor (\cref{sec:expert_data_augmentation}) & 0.1 \\
        \midrule
        \midrule
            \textit{Value Penalization (VP)} & \\
            $\lambda$ & 10 \\
            $Q^\pi_{\text{max}}$, $Q^\pi_{\text{min}}$ num. filter points (\cref{sec:maintaining_qmax_qmin_estimates}) & 50 \\
        \midrule
        \midrule
		\textit{Auxiliary Control (ACE) Scheduler} (\cref{sec:scheduler_choices}) & \\
            Num. Periods & 8 (Panda), 5 (Sawyer/Adroit) \\
            Main Task Rate & 0.5 (Panda), 0.0 (Sawyer/Adroit) \\
            Handcraft Rate & 0.5 (Panda), 1.0 (Sawyer/Adroit) \\
        \bottomrule
    \end{tabularx}
    \label{tab:hyperparameters_all}
\end{table}

Whenever possible, all algorithms and baselines use the same modifications.
\cref{tab:hyperparameters_all} also shows our choices for common off-policy RL hyperparameters as well as choices for those introduced by this work.

\textbf{DAC reward function}: for VPACE-DAC and VP-DAC, although there are many options for reward functions that map $D$ to $\hat{R}$ \citeapp{ghasemipourDivergenceMinimizationPerspective2019}, following \cite{ablettLearningGuidedPlay2023,kostrikovDiscriminatorActorCriticAddressingSample2019,fuLearningRobustRewards2018}, we set the reward to $\hat{R}_\tasks(s) = \log \left( D_\tasks(s) \right) - \log \left( 1 - D_\tasks(s) \right)$.

\textbf{$n$-step returns and entropy in TD error}: following results from \cite{eysenbachReplacingRewardsExamples2021}, we also add $n$-step returns and remove the entropy bonus in the calculation of the TD error for all algorithms in all Sawyer and Adroit environments, finding a significant performance gain for doing so.

\textbf{Absorbing states and terminal states}: for all algorithms, we do not include absorbing states (introduced in \cite{kostrikovDiscriminatorActorCriticAddressingSample2019}) or terminal markers (sometimes referred to as ``done"), since we found that both of these additions cause major bootstrapping problems when environments only terminate on timeouts, and timeouts do not necessarily indicate failure.
Previous work supports bootstrapping on terminal states when they are caused by non-failure timeouts \citeapp{pardoTimeLimitsReinforcement2018}.

\textbf{SQIL labels for policy data}: The original implementation of SQIL uses labels of 0 and 1 TD updates in \cref{eq:sqil_policy_bellman} and \cref{eq:sqil_expert_bellman}, respectively.
We found that changing the label for \cref{eq:sqil_policy_bellman} from 0 to -1 improved performance.

\textbf{Reward Scaling of .1}: we use a reward scaling parameter of .1 for all implementations.
Coupled with a discount rate $\gamma = 0.99$ (common for much work in RL), this sets the expected minimum and maximum Q values for SQIL to $\frac{-.1}{1 - \gamma} = -10$ and $\frac{.1}{1 - \gamma} = 10$.

\textbf{No multitask weight sharing}: intuitively, one may expect weight sharing to be helpful for multitask implementations.
We found that it substantially hurt performance, so all of our multitask methods do not share weights between tasks or between actor and critic.
However, the multitask discriminator in VPACE-DAC \textit{does} have an initial set of shared weights due to its significantly poorer performance without this choice.

\textbf{$\buffer^*$ sampling for $Q$}: in SQIL, DAC and RCE, we sample from both $\buffer$ and $\buffer^*$ for $Q$ updates, but not for $\pi$ updates (which only samples from $\buffer$).
The original DAC implementation in \cite{kostrikovDiscriminatorActorCriticAddressingSample2019} only samples $\buffer^*$ for updating $D$, sampling only from $\buffer$ for updating Q.

All other architecture details, including neural network parameters, are the same as \cite{ablettLearningGuidedPlay2023}, which our own implementations are built on top of.
Our code is built on top of the code from \cite{ablettLearningGuidedPlay2023}, which was originally built using \cite{rl_sandbox}.

\subsubsection{Maintaining \texorpdfstring{$Q^\pi_{\text{max}}$}{Q-pi-max}, \texorpdfstring{$Q^\pi_{\text{min}}$}{Q-pi-min} Estimates for Value Penalization}
\label{sec:maintaining_qmax_qmin_estimates}
Our approach to value penalization requires maintaining estimates for or choosing $Q^\pi_{\text{max}}$ and $Q^\pi_{\text{min}}$.
In both DAC and SQIL, the estimate of $Q^\pi_{\text{max}}$ comes from taking the mini-batch of data from $\buffer^*$, passing it through the Q function, taking the mean, and then using a median moving average filter to maintain an estimate.
The ``$Q^\pi_{\text{max}}$, $Q^\pi_{\text{min}}$ num. filter points" value from \cref{tab:hyperparameters_all} refers to the size of this filter.
We chose 50 and used it for all of our experiments.
We set $Q^\pi_{\text{min}}$ to
\begin{equation}
    Q^\pi_{\text{min}} = \frac{\text{rew. scale} \times \min(\hat{R}(s))}{1 - \gamma},
\end{equation}
where in SQIL, $\min(\hat{R}(s)) = \hat{R}(s)$ is set to 0 or -1, and in DAC, we maintain an estimate of the minimum learned reward $\min(\hat{R})$ using a median moving average filter with the same length as the one used for $Q^\pi_{\text{max}}$ .

\subsubsection{Scheduler Choices}
\label{sec:scheduler_choices}

Our \textit{ACE} algorithms use the same approach to scheduling from \cite{ablettLearningGuidedPlay2023}.
Specifically, we use a weighted random scheduler (WRS) combined a small set of handcrafted high-level trajectories.
The WRS forms a prior categorical distribution over the set of tasks, with a higher probability mass $p_{\tasks_{\text{main}}}$ (Main Task Rate in \cref{tab:hyperparameters_all}) for the main task and $\frac{p_{\tasks_{\text{main}}}}{K}$ for all other tasks.
Additionally, we choose whether to uniformly sample from a small set of handcrafted high-level trajectories, instead of from the WRS, at the \textit{Handcraft Rate} from \cref{tab:hyperparameters_all}.

Our selections for handcrafted trajectories are quite simple, and reusable between main tasks within each environment.
In the Panda tasks, there are eight scheduler periods per episode and four auxiliary tasks (\textit{reach}, \textit{grasp}, \textit{lift}, \textit{release}), and the handcrafted trajectory options are:

\begin{figure}
    \centering
    \includegraphics[width=.8\linewidth]{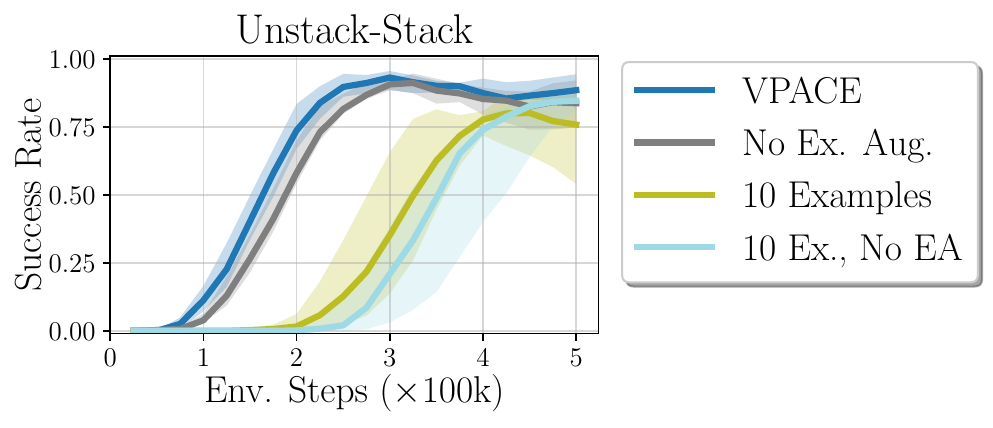}
    \caption{
        Variations of example augmentation.
        Our example augmentation scheme provides a small but noticeable bump in performance, which is magnified with when fewer expert examples are used.
    }
    \label{fig:abl_exaug}
\end{figure}

\begin{enumerate}
    \item \textit{reach}, \textit{lift}, \textit{main}, \textit{release}, \textit{reach}, \textit{lift}, \textit{main}, \textit{release}
    \item \textit{lift}, \textit{main}, \textit{release}, \textit{lift}, \textit{main}, \textit{release}, \textit{lift}, \textit{main}
    \item \textit{main}, \textit{release}, \textit{main}, \textit{release}, \textit{main}, \textit{release}, \textit{main}, \textit{release}
\end{enumerate}

In the Sawyer and Adroit environments, we actually found that the WRS was unnecessary to efficiently learn the main task, and simply used two handcrafted high-level trajectories.
In these environments, there are five scheduler periods per episode and two auxiliary tasks (\textit{reach}, \textit{grasp}), and the handcrafted trajectory options are:
\begin{enumerate}
    \item \textit{reach}, \textit{grasp}, \textit{main}, \textit{main}, \textit{main}
    \item \textit{main}, \textit{main}, \textit{main}, \textit{main}, \textit{main}
\end{enumerate}

\subsubsection[Expert Data Augmentation]{Expert Data Augmentation}
\label{sec:expert_data_augmentation}

We added a method for augmenting our expert data to artificially increase dataset size.
The approach is similar to other approaches that simply add Gaussian or uniform noise to data in the buffer \citeapp{yaratsMasteringVisualContinuous2022,sinhaS4RLSurprisinglySimple2021}.
In our case, we go one step further than the approach from \citeapp{sinhaS4RLSurprisinglySimple2021}, and first calculate the per-dimension standard deviation of each observation in $\buffer^*$, scaling the Gaussian noise added to each dimension of each example based on the dimension's standard deviation.
For example, if a dimension in $\buffer^*$ has zero standard deviation (e.g., in \texttt{Insert}, the pose of the blue block is always the same), it will have no noise added by our augmentation approach.
The parameter ``Expert Aug. Factor" from \cref{tab:hyperparameters_all} controls the magnitude of this noise, after our per-dimension normalization scheme.

In \cref{fig:abl_exaug}, we show the results of excluding expert augmentation, where there is a clear, if slight, performance decrease when it is excluded, which is even more pronounced with a smaller $\buffer^*_\tasks$ size.
All methods and baselines from our own implementation use expert data augmentation.

\subsubsection{Other Ablation Details}
\label{sec:other_ablation_details}

In our ablation experiments from, we included three baselines with full trajectory data, in addition to success examples.
We added 200 $(s,a)$ pairs from full expert trajectories to make datasets comparable to the datasets from \cite{ablettLearningGuidedPlay2023}, where they used 800 expert $(s,a)$ pairs, but their environment was run at 20Hz instead of 5Hz, meaning they needed four times more data to have roughly the same total number of expert trajectories.
We generated these trajectories using high-performing policies from our main experiments, since the raw trajectory data from \cite{ablettLearningGuidedPlay2023} would not apply given that we changed the control rate from 20Hz to 5Hz.

\subsubsection{Real Panda Implementation Details}

\begin{figure*}[t]
    \centering
    \includegraphics[width=\textwidth]{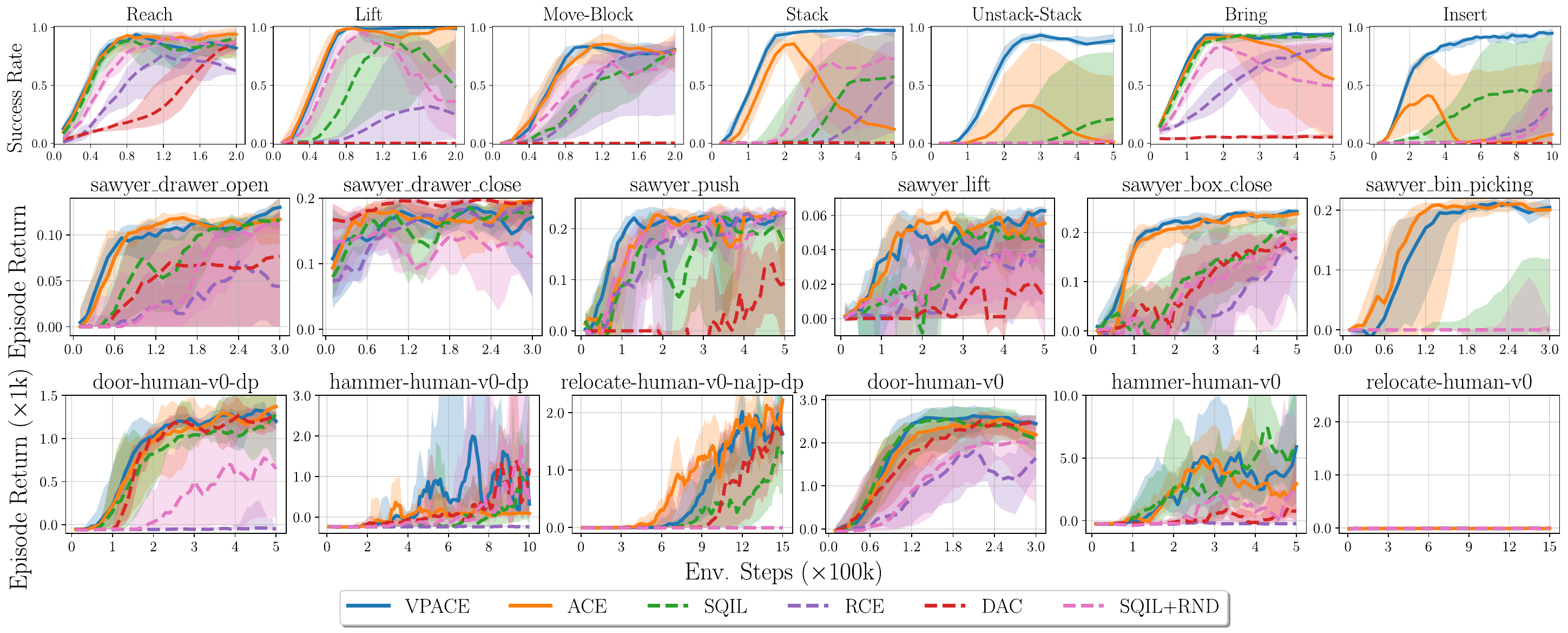}
    \caption{
        Sample efficiency performance plots for main task only for all simulated main tasks, including baselines not shown in the main text.
        Methods introduced in this work have solid lines, while baselines are shown with dashed lines.
        Performance is an interquartile mean (IQM) across 5 timesteps and 5 seeds with shaded regions showing 95\% stratified bootstrap confidence intervals \cite{agarwalDeepReinforcementLearning2021}.
    }
    \label{fig:perf_results_all_summary_more_baselines}
    \vspace{-10pt}
\end{figure*}

\begin{figure*}
    \centering
    \includegraphics[width=\linewidth]{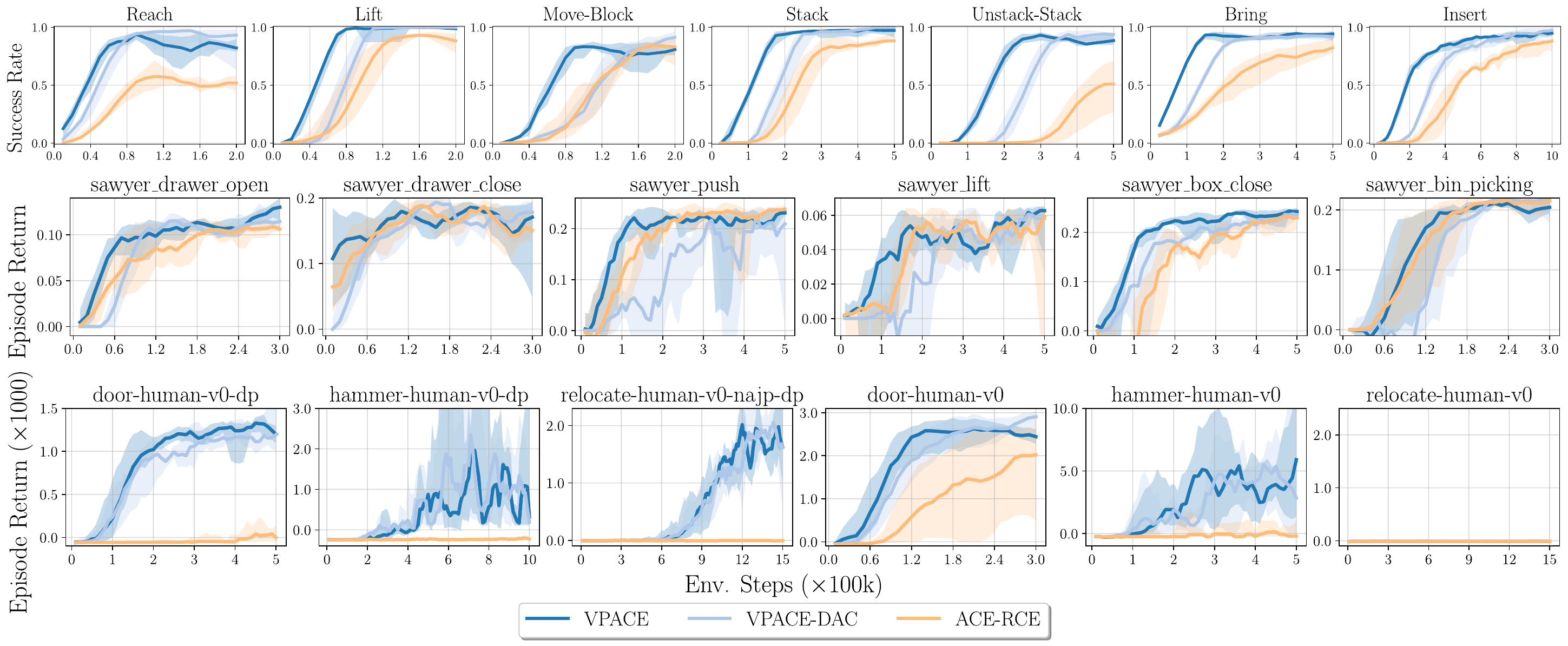}
    \caption{
        Comparison of underlying reward models when used with ACE.
        Our main results (VPACE) used SQIL as the reward model.
        VPACE-DAC eventually reaches similar performance to VPACE-SQIL, but tends to take longer, and ACE-RCE often has far poorer performance than both.
    }
    \label{fig:ace_var_all}
    \vspace{-10pt}
\end{figure*}

\begin{figure*}
    \centering
    \includegraphics[width=\linewidth]{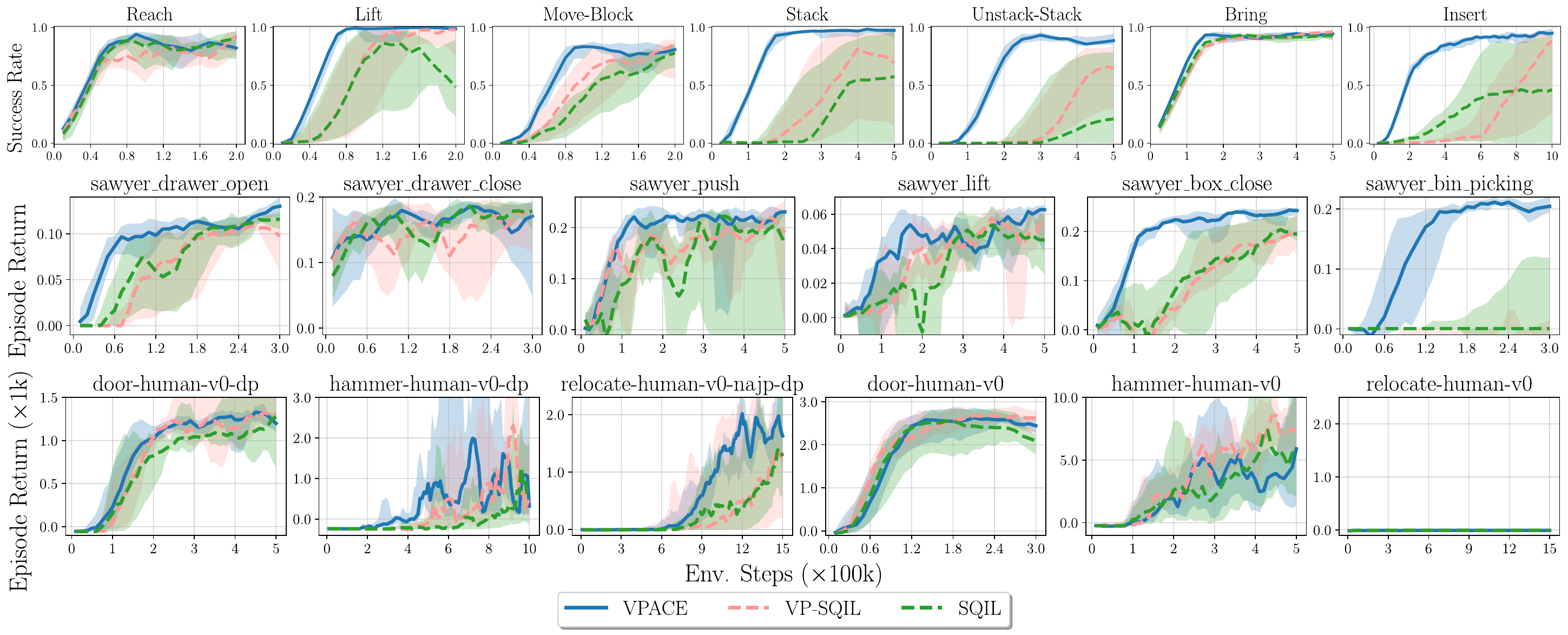}
    \caption{
        Comparison of different variants of SQIL: both VP and ACE (VPACE), VP-only (VP-SQIL), or SQIL alone.
        VP-SQIL generally either results in an improvement or has no effect on performance, compared with SQIL, but the exclusion of ACE results in far poorer performance than VPACE, especially for the most complex tasks.
    }
    \label{fig:vp_var_all}
    \vspace{-10pt}
\end{figure*}

\begin{figure*}
    \centering
    \includegraphics[width=\linewidth]{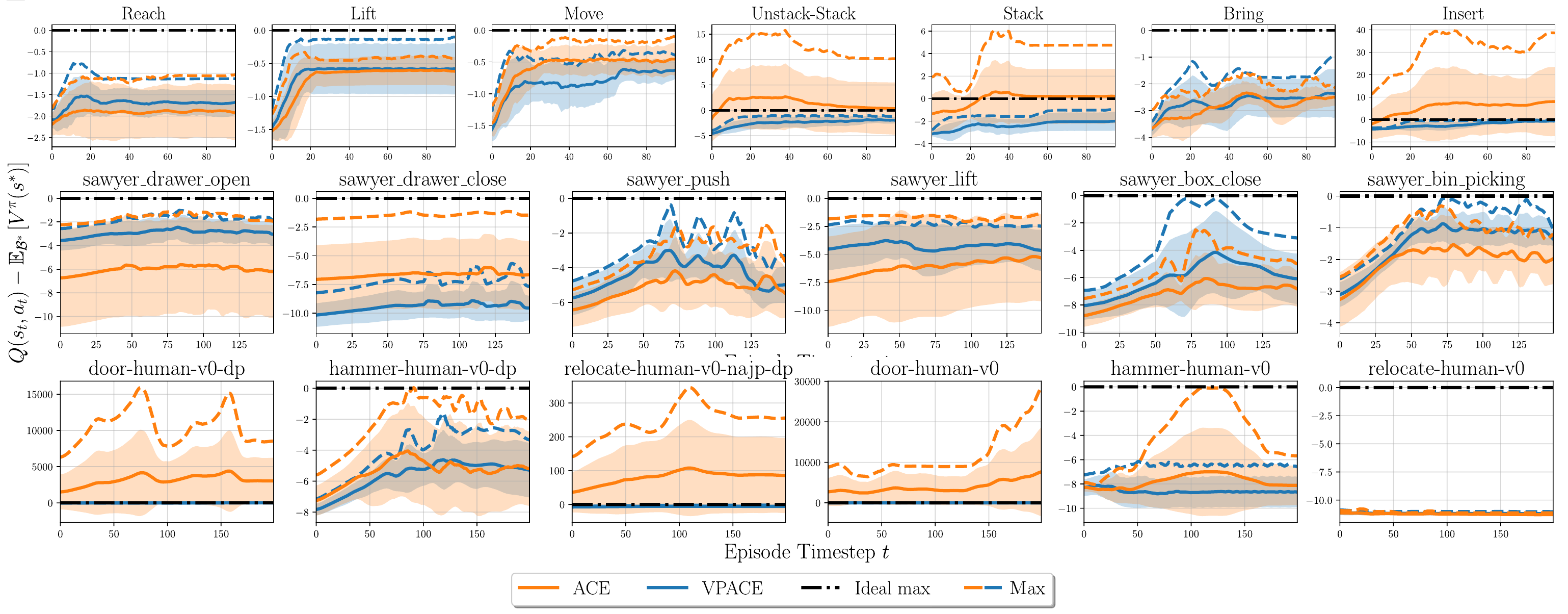}
    \caption{
        Comparison of different variants of SQIL: both VP and ACE (VPACE), VP-only (VP-SQIL), or SQIL alone.
        VP-SQIL generally either results in an improvement or has no effect on performance, compared with SQIL, but the exclusion of ACE results in far poorer performance than VPACE, especially for the most complex tasks.
    }
    \label{fig:qpert_all}
    \vspace{-10pt}
\end{figure*}

While most of the design choices in \cref{sec:additional_impl_details} apply to all environments tested, our real Panda environment had some small specific differences, mostly due to the complications of running reinforcement learning in the real world.
We list the differences here, but for an exhaustive list, our open source code contains further details.

\textbf{Maximum episode length: } The maximum episode length for both \texttt{RealDrawer} and \texttt{RealDoor} is 1000 steps, or 200 seconds in real time.
This was selected to reduce how often the environment had to be reset, which is time consuming.
Running episodes for this long, and executing actions at 5 Hz, our environments complete 5000 environment steps in roughly 20 minutes.
The extra time is due to the time to reset the environment after 1000 steps or after a collision.
VPACE took approximately 100 minutes to learn to complete \texttt{RealDrawer} consistently, and about 200 minutes to learn to complete the more difficult \texttt{RealDoor}.

\textbf{Shorter initial exploration: } To attempt to learn the tasks with fewer environment samples, we reduce buffer warmup to 500 steps, and initial random exploration to 1000 steps.

\textbf{Frame stack: } For training, we stacked two regular observations to avoid state aliasing.

\textbf{Ending on success: } We ended episodes early if they were determined to be successful at the main task only.
Although this is not necessary for tasks to be learned (and this information was \textit{not} provided to the learning algorithm), it gave us a way to evaluate training progress.

\textbf{Extra gradient steps: } To add efficiency during execution and training, we completed training steps during the gap in time between an action being executed and an observation being gathered.
Instead of completing a single gradient step at this time, as is the case for standard SAC (and VPACE), we completed four gradient steps, finding in simulated tasks that this gave a benefit to learning efficiency without harming performance.
Previous work \cite{ablettLearningGuidedPlay2023}, \citeapp{chenRandomizedEnsembledDouble2021} has found that increasing this update rate can reduce performance, but we hypothesize that our value penalization scheme helps mitigate this issue.

\textbf{Collisions: } If the robot is detected to have exceeded force or torque limits (20 N and 20 Nm in our case, respectively), the final observation prior to collision is recorded, and the environment is immediately reset.
There are likely more efficient ways to handle such behaviour, but we did not investigate any in this work.

\section{Additional Performance Results}
\label{sec:add_performance_results}

In this section, we expand upon our performance results and our Q-value overestimation analysis.

\subsection{Expanded Main Task Performance Results}
\label{sec:expanded_main_task_perf_results}

\cref{fig:perf_results_all_summary_more_baselines} shows expanded results for our simulated environments, with each baseline shown for each individual environment.

\subsection{ACE Reward Model Comparison}
\label{subsec:ace_reward_model_comparison}

The SAC-X framework, which we describe as ACE when used with example states only, is agnostic to the choice of reward model (see \cref{app:reward_model_options}).
Therefore, we also completed experiments in each of our simulated environments with DAC and with RCE as the base reward model, instead of SQIL, which was used for all of our main VPACE results.
The results are shown in \cref{fig:ace_var_all}, where it is clear that VPACE with SQIL learns more efficiently than VPACE with DAC and with higher final performance than ACE with RCE.
We do not use value penalization (VP) for RCE, since RCE uses a classification loss for training Q that would not benefit from the use of value penalization.

\subsection{Single-task Reward Model Comparison}
\label{subsec:st_reward_model_comparison}

\begin{figure}
    \centering
    \includegraphics[width=.9\linewidth]{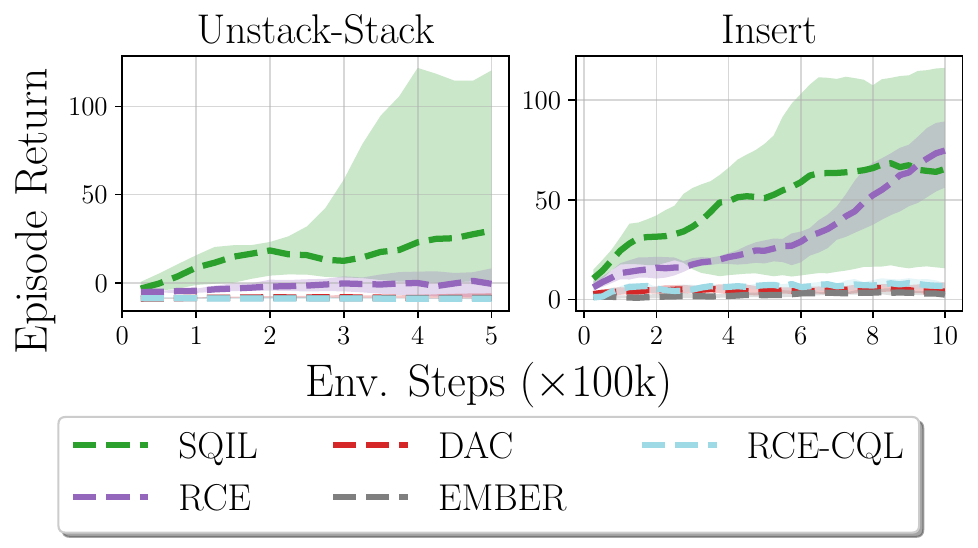}
        \caption{
            Results for different reward models for EBC, excluding the use of both value penalization (VP) and auxiliary control from examples (ACE).
        }
        \label{fig:perf_results_abl_rew_model}
    \vspace{-12pt}
\end{figure}

We compare SQIL, DAC, and RCE , without adding value penalization (VP) or auxiliary control from examples (ACE) in \cref{fig:perf_results_abl_rew_model}, with two additional, more recent EBC baselines: \textbf{EMBER} \cite{wuExampleDrivenModelBasedReinforcement2021} and RCE combined with conservative Q-learning (\textbf{RCE-CQL})\cite{kumarConservativeQLearningOffline2020}, \citeapp{hatchExampleBasedOfflineReinforcement2021}).
We find that both EMBER and RCE-CQL perform quite poorly, and much more poorly than SQIL, DAC, and RCE, further justifying their use as our primary reward models in our main experiments.

\subsection{Single-task Value Penalization}
\label{subsec:single_task_value_penalization}

Our scheme for value penalization, while initially motivated by the use of ACE, can be used in the single task regime.
In \cref{fig:vp_var_all}, we compare the performance of VPACE, SQIL, and SQIL with value penalization, but \textit{without} auxiliary tasks (VP-SQIL).
VP-SQIL either has no effect on performance or results in an improvement over SQIL, as expected, but is still strongly outperformed by VPACE on the more complicated tasks, such as \texttt{Stack}, \texttt{Unstack-Stack}, \texttt{Insert}, \texttt{sawyer\_box\_close}, \texttt{sawyer\_bin\_picking}, and the \texttt{dp} variant of \texttt{relocate-human-v0}.

\subsection{Q-Value Overestimation and Penalization -- All Environments}
\label{sec:add_value_pen_benefits_results}

\cref{fig:qpert_all} shows the same value overestimation analysis plots shown in the original paper for all tasks in.
The results for other difficult tasks also show clear violations of $y(s,a) \leq y(s^*,a^*)$, and tasks in which this rule is violated also often have poorer performance.
Intriguingly, although the rule is severely violated for many Adroit hand tasks, ACE without VP still has reasonable performance in some cases.
This shows that highly uncalibrated Q estimates can still, sometimes, lead to adequate performance.
We hypothesize that this occurs because these tasks do not necessarily need $s_T \sim \buffer$ to match $s^* \sim \buffer^*_{\text{main}}$ to achieve high return, but we leave investigating this point to future work.

\section{Why Does SQIL Outperform RCE?}
\label{sec:sqil_vs_rce}

\begin{figure*}
    \centering
    \includegraphics[width=.7\linewidth]{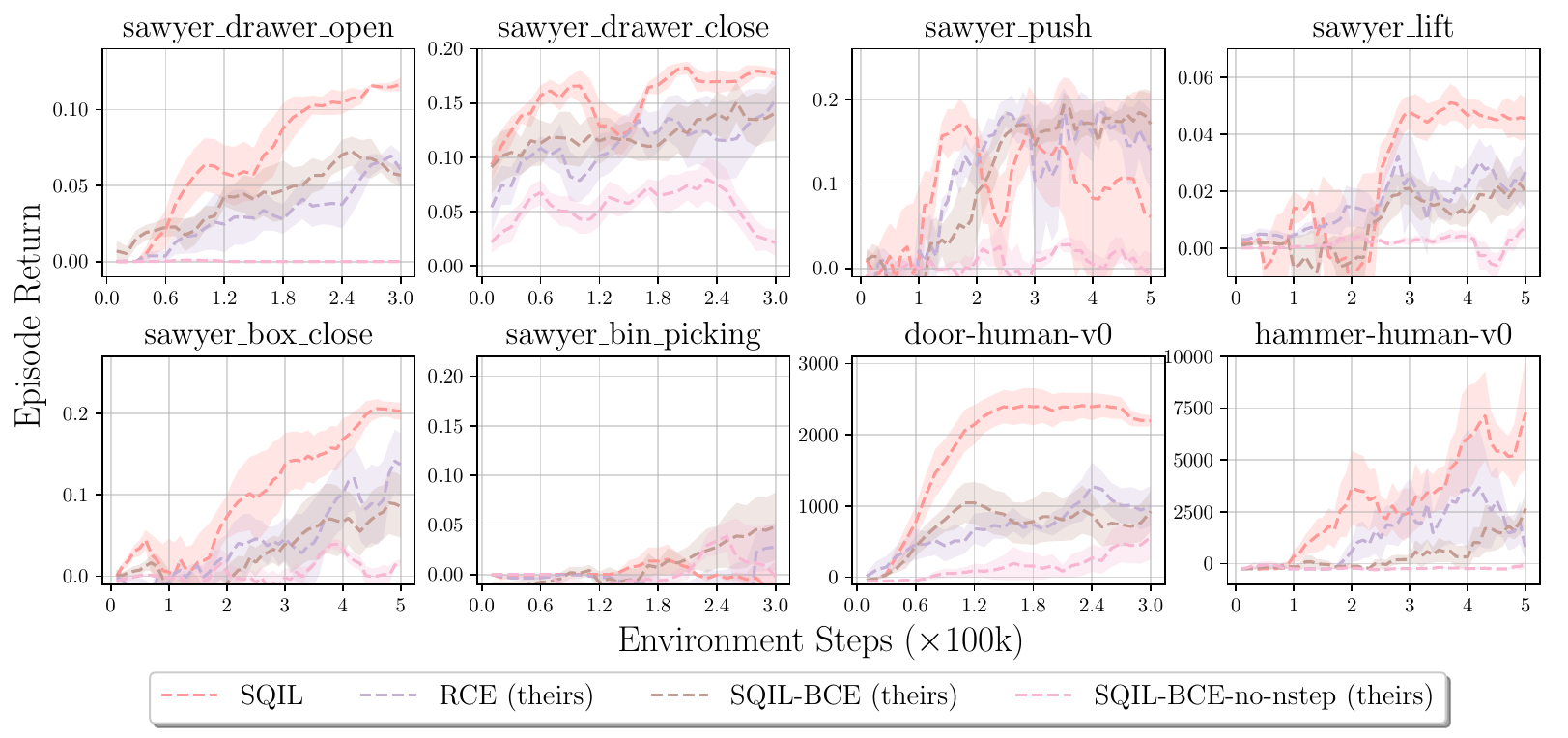}
    \caption{
        Performance results on the Sawyer and Adroit tasks considered in \cite{eysenbachReplacingRewardsExamples2021}, using their own implementations based on binary cross entropy (BCE) loss, but with an additional SQIL-BCE (including $n$-step) baseline.
        We also show our implementation of SQIL (without value penalization or auxiliary tasks), with Mean Squared Error TD updates, which outperforms all of the BCE-based methods on average.
    }
    \label{fig:rce_hand_theirs_performance}
\end{figure*}

Our results show that VPACE-SQIL outperforms ACE-RCE, and that VP-SQIL and SQIL outperform RCE in almost all cases.
This result is in conflict with results from \cite{eysenbachReplacingRewardsExamples2021}, which showed SQIL strongly outperformed by RCE.
In this section, we show results that help explain our findings.

\cite{eysenbachReplacingRewardsExamples2021} claimed that the highest performing baseline against RCE was SQIL, but it is worth noting that their implementation\footnote{Available at \url{https://github.com/google-research/google-research/tree/master/rce} at time of writing.} is a departure from the original SQIL implementation, which uses MSE for TD updates, described in \cref{sec:reward_labels_regression}.
Furthermore, \cite{eysenbachReplacingRewardsExamples2021} also noted that it was necessary to add $n$-step returns to off-policy learning to get reasonable performance.
In their experiments, however, this adjustment was not included for their version of SQIL with BCE loss.
We complete the experiments using their implementation and show the average results across all RCE environments in \cref{fig:rce_hand_theirs_performance}, and also compare to using SQIL with MSE (without value penalization or auxiliary tasks) for TD updates.

These results empirically show that RCE and SQIL with BCE loss perform nearly identically, indicating that benefits of the changed TD targets and weights described in \cref{sec:reward_labels_classification} may not be as clear as previously described.
SQIL with MSE clearly performs better on average, although it still performs worse than VPACE (see \cref{sec:add_performance_results}).

\section{Limitations}
\label{sec:limitations}

In this section, we detail limitations of this work.

\textbf{Experimental limitation---Numerical state data only} All of our tests are done with numerical data, instead of image-based data.
Other work \cite{riedmillerLearningPlayingSolving2018}, \citeapp{yaratsMasteringVisualContinuous2022} has shown that for some environments, image-based learning just results in slowing learning compared with numerical state data, and we assume that the same would be true for our method as well.

\textbf{Assumption---Generating example success states is easier} We claim that success example distributions are easier to generate than full trajectory expert data, and while we expect this to be true in almost all cases, there may still be tasks or environments where accomplishing this is not trivial. As well, similar to other imitation learning methods, knowing how much data is required to generate an effective policy is unknown, but adding a way to append to the existing success state distribution (e.g., \cite{singhEndtoEndRoboticReinforcement2019}) would presumably help mitigate this.

\textbf{Assumption/failure mode---Unimodal example distributions} Although we do not explicitly claim that unimodal example state distributions are required for VPACE to work, all of our tested tasks have roughly unimodal example state distributions.
It is not clear whether our method would gracefully extend to the multimodal case, and investigating this is an interesting direction for future work.

\textbf{Experimental limitation---Some environment-specific hyperparameters} While the vast majority of hyperparameters were transferable between all environments and algorithms, the scheduler period, the inclusion of $n$-step targets, and the use of entropy in TD updates, were different between environments to maximize performance.
Scheduler periods will be different for all environments, but future work should further investigate why $n$-step targets and the inclusion of entropy in TD updates makes environment-specific differences.

\subsection{VPACE and LfGP Limitations}
VPACE shares the following limitations with LfGP \cite{ablettLearningGuidedPlay2023}.

\textbf{Assumption---Existence of auxiliary task datasets} VPACE and LfGP require the existence of auxiliary task example datasets $\buffer^*_\text{aux}$, in addition to a main task dataset $\buffer^*_\text{main}$.
This places higher initial burden on the practitioner.
In future, choosing environments where this data can be reused as much as possible will reduce this burden.

\textbf{Assumption---Clear auxiliary task definitions} VPACE and LfGP require a practitioner to manually define auxiliary tasks.
We expect this to be comparatively easier than generating a similar dense reward function, since it does not require evaluating the relative contribution of individual auxiliary tasks.
As well, all tasks studied in this work share task definitions, and the panda environment even shares task data itself, leading us to assume that these task definitions will extend to other manipulation tasks as well.

\textbf{Assumption---Clear choices for handcrafted scheduler trajectories} VPACE and LfGP use a combination of a weighted random scheduler with a handcrafted scheduler, randomly sampling from pre-defined trajectories of high level tasks.
\cite{ablettLearningGuidedPlay2023} found that the handcrafted scheduler added little benefit compared with a weighted random scheduler, and further work should investigate this claim, or perhaps attempt to use a learned scheduler, as in \cite{riedmillerLearningPlayingSolving2018}.

\subsection{Reinforcement Learning Limitations}

\textbf{Experimental limitation---Free environment exploration} As is common in reinforcement learning methods, our method requires exploration of environments for a considerable amount of time (on the order of hours), which may be unacceptable for tasks with, e.g., delicate objects.

\bibliographystyleapp{IEEEtran}
\bibliographyapp{IEEEabrv, vpace}

\end{document}